\documentclass{article}

\newcommand{\dissertationtitle}{LSTM-UT: Bounded Gated Memory for Looped Transformers}

\usepackage{iclr2027/iclr2027_conference,times}
\usepackage{graphicx}
\graphicspath{{dis-images/}}
\usepackage{amssymb}
\usepackage{amsmath}
\usepackage{microtype}
\usepackage{booktabs}
\usepackage{array}
\usepackage{multirow}
\usepackage{float}
\usepackage{placeins}
\usepackage{url}
\usepackage[hidelinks]{hyperref}
\hypersetup{
  pdftitle={\dissertationtitle},
  pdfauthor={Aras Kavuncu, Muhammad Burhan Hafez}
}

\newcolumntype{L}[1]{>{\raggedright\arraybackslash}p{#1}}
\newcolumntype{C}[1]{>{\centering\arraybackslash}p{#1}}
\newcolumntype{R}[1]{>{\raggedleft\arraybackslash}p{#1}}

\title{\dissertationtitle}
\author{Aras Kavuncu \\
University of Southampton \\
\texttt{tak1e25@soton.ac.uk} \\
\texttt{aras.kavuncu@yahoo.com}
\And
Muhammad Burhan Hafez \\
University of Southampton \\
\texttt{burhan.hafez@soton.ac.uk}}
\iclrfinalcopy
\begin{document}
\maketitle
\lhead{Preprint} 

\begin{abstract}
Can the loop of a looped Transformer serve as a rollout, with one application of the shared block per step of a dynamical system? We started 
with two existing designs, one that carries only its residual stream and one, CoTFormer, that also caches every earlier state. On Rule 30, a 
chaotic cellular automaton, both lose the computation shortly after their trained depth. LSTM-UT, which adds an LSTM-style gated cell, does not, and it recalls earlier
states better. Taking it apart shows that each ability has a simpler source. For rollouts it is decay: a fixed decay on the plain model's stream is
sufficient. For recall it is an input-dependent gate: at the longest horizon a gated stream with no cell recalls as well as LSTM-UT, and no
fixed decay helps. The two do not simply add, since a gated stream with a fixed decay does neither reliably. Storing every earlier state hurts extrapolation, where CoTFormer fits and rolls out further with its cache cut to the most recent
repeat; in recall, the stored copy is looked up at the shortest horizon and not at the next. LSTM-UT, whose forget gate makes the decay itself input-dependent, is the only model in the leading group on both. The forget gate was added to the LSTM so that a state carried at weight one could be
reset instead of accumulating; we find the same need over depth.
In language modelling LSTM-UT is competitive with CoTFormer at lower training cost.
Code is available at \url{https://github.com/arskvnc22/lstm-ut}.
\end{abstract}

\section{Introduction}

Environment dynamics are an iterative process: a state evolves by repeated application of an approximately stationary transition law.
A looped Transformer has the same structure. It applies one shared block repeatedly to a latent state, so the number of steps is set at
inference time and needs no new parameters \citep{dehghani2018universal}; we call the plain version, with nothing carried but its
residual stream, the Block Universal Transformer (BUT). This makes looping a natural backbone
for learned simulators and world models. Dreamer already rolls out a recurrent world model, a recurrent state updated once per step
\citep{hafner2020dreamer,hafner2025dreamerv3}, and recent work builds world models directly on looped Transformers: LoopWM refines each
latent transition with a shared block \citep{lu2026loopwm}, and Test-Time Forecasting (TTF) uses
recurrent depth to forecast unseen ODE and PDE dynamics in context after training only for one-step prediction \citep{lai2026testtime}.
In both, the loop refines a single transition and the rollout over time is a separate outer loop. We ask instead whether the loop
itself can be the rollout, with one application of the shared block per step of the dynamics.

Using the loop as the rollout asks for two things that work on looped transformers has rarely tested. The first is a
stable rollout. Most looped models are built to reach a fixed point, where further iterations leave a solved state
unchanged \citep{bansal2022overthinking,movahedi2026fixedpoint,huang2026equilibrium}, and existing stabilisers serve that goal:
they keep the looped state bounded \citep{lu2026loopwm,prairie2026parcae} or drive it to a fixed point
\citep{movahedi2026fixedpoint,manek2019stable}. In a rollout every iteration must advance the state, and a model whose state
settles has failed. Learned simulators face the same problem \citep{sanchezgonzalez2020gns,lippe2023pderefiner,schiff2024dyslim},
because small errors compound over long horizons \citep{talvitie2017selfcorrecting,xiao2019compounding,lambert2022compounding}.
What a rollout needs is for errors to shrink while the state keeps moving: contraction towards the set of valid states, not
towards a single point.

The second is memory of earlier states. World models carry a recurrent state so that information survives many rollout steps \citep{hafner2019planet}, and how long it survives limits them \citep{deng2023s4wm}.
Contraction alone does not provide this. A map that contracts everywhere has a single attractor and can latch nothing;
\citet{bengio1994learning} showed that a state is stored robustly where the dynamics contract around one of several attractors, and
that gradients to earlier steps vanish there; the LSTM answered with a path whose retention is gated \citep{hochreiter1997lstm,gers2000forget}.
For a loop, a contraction applied uniformly protects the current state but cannot single out what should outlast it.
Chain-of-thought \citep{wei2022cot} suggests a different route, iterative refinement and rereading earlier results, and CoTFormer brings this into a
looped Transformer through depth-wise attention: each repeat attends to the cached keys and values of every earlier repeat, so
earlier states stay available instead of having to survive in the current one \citep{mohtashami2025cotformer}.

We study both requirements on Rule~30, a chaotic cellular automaton whose correct state is known at every step. We tie the number
of applications of the shared block to the number of automaton steps: a model run for \(R\) repeats is trained to produce the row
\(R\) steps ahead, so running the loop past its trained depth is a rollout past its trained horizon. This gives each loop step a
meaning, unlike Loop-OWM, whose fixed number of steps towards a single output leaves an individual step undetermined
\citep{gao2026slots}. \emph{Depth extrapolation} trains on depths \(1,\dots,K\), supervises only the final repeat, and asks
whether the transition keeps advancing the automaton correctly beyond \(K\). Within the training range, many learned
transitions fit equally well and held-out accuracy cannot tell them apart \citep{damour2022underspecification}:
Transformers trained on automata can find shortcuts that are exact in-distribution but fail under shift
\citep{liu2023shortcuts}, and models that predict trajectories well need not have learned the law that generates
them \citep{vafa2025foundation}. Because Rule~30 is chaotic, a transition that only approximates the rule drifts
quickly past \(K\). \emph{Delayed recall} instead supervises every repeat to decode to its automaton row, so the state to be
recalled is known; the model must then keep computing and recover it, a test akin to the context-dependent recall of \citet{deng2023s4wm}.

We began with BUT and CoTFormer. Both lose the computation soon after their trained depth: BUT within 19 steps, CoTFormer within 3,
or 8 with its cache cut to the most recent repeat. LSTM-UT, which we built in response, adds a bounded, gated cell and keeps going.
Taking it apart showed that each of its two abilities has a simpler source. \emph{A decaying state gives stable rollouts}: every
model but one with a decaying state, in its stream or in a side state read back into the stream, rolls out far past its trained depth, and
those without one do not reliably, whatever gating, re-injection, width or attention over earlier repeats they add. Decay makes
the rollout contract, so small errors die out and the model usually resumes a valid computation after large ones. \emph{Input-dependent
gates help recall}: at the longest horizon a gate on an undecayed stream lengthens the span of rows a model can recall, no fixed decay does,
and CoTFormer's stored history is looked up at horizon 6 and not at horizon 12. \emph{The two do not add}: gated BUT with a fixed decay,
the exception above, also recalls no better than BUT. LSTM-UT, whose forget gate makes the decay input-dependent, is the only model
in the leading group on both tasks, and it reads some recalled rows from its cell. In language modelling it is competitive with CoTFormer at lower
cost.
Our contributions are:
\begin{itemize}
\item \textbf{Decay gives stable rollouts.}  Trained on depths 1-12, six model families with a decaying state are exact at depth 60 in all 30 seeds, and 29 of those seeds 
still have more than half their rows exact at depth 1,000. Three families with neither decay nor a side state stop between depths 18 and 31 in all 15 seeds. A fixed decay of 
0.9 on BUT's stream, with nothing else changed, is sufficient. The exception is decaying gated BUT, where the same decay reaches depth 60 in two of five seeds.
\item \textbf{Contraction and recovery.} The six decaying families halve a small error in the stream in at most 6 repeats (gated decay BUT in 7), the undecayed-stream models in at least
32. After damage that flips bits, \(52\)--\(94\%\) of branches return to a faithful Rule~30 computation (\(39\%\) for gated decay BUT), against at most \(10\%\).
\item \textbf{Input-dependent gates help recall; no fixed decay does, and one undoes the gate.} After 25,000 training steps at horizon 20, gated
BUT and LSTM-UT recall \(70.5\%\) and \(67.6\%\) of rows exactly and BUT \(53.8\%\). The four ungated fixed-decay models recall
\(53.7\%\) and \(44.7\%\) (decay BUT with \(\lambda=0.9\) and \(0.5\)), \(49.7\%\) (linear memory) and \(40.0\%\) (FPRM BUT). Gated BUT with a decay of 0.9 recalls \(53.0\%\).
\item \textbf{Storing does not guarantee use.} CoTFormer's expanding cache can act as addressable memory: at horizon 6, two of
three models fall to chance when the cached copy of the requested row is given another input's values. At horizon 12 no
fitted model depends on that copy, and at horizon 20 no CoTFormer seed fits its rollout. On Rule~30, CoTFormer fits reliably and extrapolates
more than three steps only when its cache is cut to the most recent repeat, and correcting a failed state rescues the rollout more robustly when the cache is also cleared
(Appendix~\ref{app:cotformer}).
\item \textbf{LSTM-UT.} A bounded, gated cell state added to a shared Transformer block, built after BUT and CoTFormer failed to
extrapolate. It is the only model in the leading group on both tasks, and its two ingredients do not reproduce it without the cell: a gated
stream with a fixed decay extrapolates in two of five seeds and recalls at BUT's level. Its cell is used as memory: after a
swap between two inputs, recalled rows follow the donor at some ages. On OpenWebText2 it is ahead of BUT and competitive with
CoTFormer, which needs up to \(91\%\) more training time per step.
\end{itemize}

\section{Background and Related Work}
\subsection{Recurrent depth and depth extrapolation}
Recurrent-depth Transformers apply shared layers repeatedly to an evolving hidden state \citep{dehghani2018universal}. More
recurrent depth at inference can improve reasoning \citep{geiping2026hugin,saunshi2025latentthoughts}, and looping has been built
into large-scale pretraining \citep{zhu2025ouro}.

Whether a learned transition keeps working beyond the trained depths has mainly been studied as easy-to-hard generalisation.
Recurrent networks trained on small problems can solve larger ones by running more iterations
\citep{schwarzschild2021easyhard}, but can degenerate when many are applied, a failure \citet{bansal2022overthinking} call
overthinking and mitigate by re-injecting the input; \citet{kohli2026loop} observe it in recurrent-depth Transformers at very deep
compositions. Stability-oriented work instead drives the recurrence towards a fixed point: STARS by regularisation
\citep{yang2026stabilizing}, and Fixed-Point Reasoners by scaling the residual stream \citep{movahedi2026fixedpoint}. LoopWM,
following Parcae, makes only the linear retention term contractive \citep{lu2026loopwm,prairie2026parcae}, which bounds the state.
A fixed point is the wrong target for a model that must keep advancing a cellular automaton, but the stabiliser carries over: 
FPRM BUT, which is BUT with their residual scaling, extrapolates to depth 60 in every seed when trained on rollouts, and contracts fastest of the models we test (Section~\ref{sec:ablations}). 
In learned simulators, long-horizon drift has been traced
to transient amplification of perturbations even when the dynamics are asymptotically stable \citep{pervez2026transient}.

\subsection{Memory across depth}
A Transformer without dedicated storage can borrow it from its own representations: vision Transformers turn background patch
tokens into scratch space unless given register tokens \citep{darcet2024registers}, and a single-block Universal Transformer fails
on Sudoku without learned memory tokens \citep{sapunov2026utmemory}. Work that treats depth as a sequence axis offers further
storage across layers \citep{heddes2025dca,kimi2026attentionresiduals,xie2025mhc}. Depth Memory casts these as storing and
retrieving information across layers \citep{gonzalezmartinez2026depthmemory}: explicit mixers attend over a cache that grows with
depth, while implicit mixers write into a fixed-size state. Its Linear Depth Memory keeps the residual stream and adds a separate
associative memory, and its ablations show that keeping the two streams separate matters.

Looped models face the same choice across recurrent depth. CoTFormer is an explicit mixer: each repeat attends to the
keys and values of every earlier repeat \citep{mohtashami2025cotformer}, which hold earlier recurrent states that the current state need not
retain. LSTM-UT is an implicit mixer: like Linear Depth Memory, it leaves the block's proposal on the residual stream
and writes to a separate fixed-size cell, whose gates decide what persists across repeats; MeSH and persistent-memory
recursive Transformers also add managed state to a loop \citep{yu2026mesh,farhan2026persistentmemory}, and gates have been
used inside Transformers before \citep{parisotto2020gtrxl,kamel2026gated}. We do not claim
that a gated cell beats depth-wise attention in fixed-depth models, where every depth is seen in training. In a loop
run past its trained depth, an explicit history keeps growing and must be read at depths never trained on, while an
implicit state stays fixed in size but must be rewritten stably by one shared rule; our experiments ask which survives.

\section{Models}
\label{sec:architecture}
The models differ in what is carried from one repeat to the next and in
whether it decays (Appendix Table~\ref{tab:models}). Let \(x\) be the stream entering a repeat, \(\bar h=\mathcal T_\theta(x)=x+\Delta(x)\)
the block's output, \(x^+\) the next stream, and \(e\) the embedded input.

\emph{BUT} carries only its stream, \(x^+=\bar h\). \emph{BUT with input injection} computes \(x^+=\mathcal T_\theta(x+e)\), the
standard remedy for depth extrapolation in looped models
\citep{bansal2022overthinking,geiping2026hugin,yang2024loopedlearning,lai2026testtime}. \emph{Decay BUT} computes
\(x^+=\lambda x+\Delta(x)\) with a fixed \(\lambda\) of 0.9 or 0.5, the leaky integrator of reservoir computing
\citep{jaeger2007leaky}; BUT is \(\lambda=1\) and nothing else differs. \emph{FPRM BUT} \citep{movahedi2026fixedpoint} learns
per-channel residual scales that keep about \(14\%\) of the stream per repeat and re-adds \(e\) every repeat; we also train it
without the re-injection. \emph{CoTFormer} is BUT whose attention also reads the cached keys and values of every earlier repeat.

BUT and CoTFormer did not extrapolate or recall reliably on Rule~30, so we built a model with a small
state that it must actively write to and read from. \emph{LSTM-UT} adds a separate, bounded cell state to the shared block, updated with
the input, forget, and output gates of an LSTM \citep{hochreiter1997lstm,gers2000forget}. A shared memory layer normalisation gives
\(\hat x=\operatorname{LN}_m(x)\) and \(h=\operatorname{LN}_m(\bar h)\). LSTM-UT updates its cell \(c\) and residual stream as
\begin{align}
f&=\sigma(W_{fx}\hat x+W_{fh}h),&
i&=\sigma(W_{ix}\hat x+W_{ih}h),\\
\widetilde c&=\tanh(W_{cx}\hat x+W_{ch}h),&
c^+&=f\odot c+i\odot\widetilde c,\\
o&=\sigma(W_{ox}\hat x+W_{oh}h),&
x^+&=\bar h+o\odot\tanh(c^+).
\end{align}
The block's output is kept and the output gate adds a read from memory, so the gates decide what persists across repeats, not how
much of the stream each repeat overwrites. The cell starts at zero for each input and is not given to attention or the MLP.

Four controls isolate parts of this design. \emph{Gated BUT} has no cell and gates the block's update instead: from the same \(\hat x\)
and \(h\) it computes \(G=\sigma(W_{gx}\hat x+W_{gh}h)\) and \(x^+=G\odot\bar h+(1-G)\odot x=x+G\odot\Delta(x)\), a highway gate
\citep{srivastava2015highway} across depth as in the Gated Recurrent Transformer \citep{hegazy2026gated}. \emph{Gated decay BUT} adds decay BUT's fixed \(\lambda=0.9\) to it: \(x^+=\lambda x+G\odot\Delta(x)\). \emph{LSTM-UT without the forget gate} uses \(c^+=c+i\odot\widetilde c\).
\emph{Linear memory} keeps the separate cell and removes everything else:
\begin{equation}
c^+=\lambda\odot c+W_x\hat x+W_hh,\qquad x^+=\bar h+c^+,
\end{equation}
where \(\lambda=\sigma(\ell)\in(0,1)\) is one learned constant per channel: no gates and no \(\tanh\), a diagonal linear
recurrence as in the Linear Recurrent Unit \citep{orvieto2023resurrecting}, run over repeats.

\section{Rule 30 depth extrapolation}
\label{sec:rule30}
\label{sec:ablations}
Depth extrapolation asks whether each further application of the shared block still performs one correct Rule~30 update beyond the
supervised range. LSTM-UT adds several things to BUT at once: a second carried state, a forget gate that decays it, gated and
saturating writes and reads, and parameters. We separate these with variants and with interventions on trained models.

\subsection{Task and protocol}
Rule~30 \citep{wolfram1986random} synchronously updates each binary cell from its two periodic neighbours: \(\rho(x)_i=x_{i-1}\oplus(x_i\lor x_{i+1})\). Models receive a
64-cell initial row and apply one shared Transformer block per requested update. All models are built on the GPT-2 block \citep{radford2019language}, 
modified to use bidirectional attention and rotary positions \citep{su2024roformer}; the principal models use width 64, 
four heads, an MLP width of 4 times the embedding
dimension, and zero dropout. Unlike recurrent-depth language models that surround the looped block with non-recurrent prelude and coda
layers \citep{geiping2026hugin}, our cellular-automaton models contain only the shared block.
We write \(R_K\) for training that cycles deterministically through depths \(1{:}1,\ldots,K{:}K\), one depth per optimizer step. Only the
final row is supervised, and no clean intermediate states are supplied. \(R_6\) uses 100,000 training rows and 7,500 steps; \(R_{12}\) uses
500,000 rows and 24,000 steps. Validation and final-test splits each contain 4,096 independently generated rows. Checkpoints are selected by
validation cell accuracy at an unseen depth (9 for \(R_6\), 15 for \(R_{12}\)), with no minimum in-distribution score, and the final-test split
is used only for reporting. We report cell accuracy and exact-row accuracy, which requires all 64 cells to be correct.

\begin{figure}[t]
\centering
\includegraphics[width=0.6\textwidth]{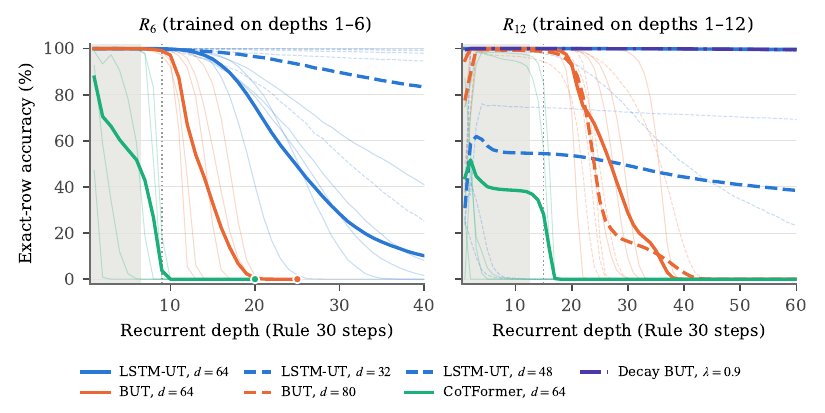}
\caption{Rule~30 depth extrapolation: final-test exact-row accuracy against recurrent depth for the validation-selected unconstrained
extrapolation checkpoints. Bold lines are means over five model-initialisation seeds; faint lines are individual seeds. Grey bands mark
the supervised depths and dotted lines the checkpoint-selection depth. Lines ending in a dot were not evaluated further; every seed was
already at \(0\%\) exact-row accuracy there. Full cell and exact-row values are in Appendix Tables~\ref{tab:rule30-r6-r8-extrapolation}
and~\ref{tab:rule30-r12-extrapolation}. Decay BUT differs from BUT only in a fixed multiplier of \(0.9\) on the stream; its line
lies on LSTM-UT's. CoTFormer fits the supervised depths in one of its five \(R_{12}\) seeds (Appendix~\ref{app:cotformer}).}
\label{fig:rule30-extrapolation}
\end{figure}

\subsection{Variants and measurements}
\label{sec:abl-measure}
All variants use the \(R_{12}\) protocol with five seeds; Appendix~\ref{app:ablations} gives per-seed results and further controls.
\paragraph{Variants.}
Table~\ref{tab:abl-main} groups the models of Section~\ref{sec:architecture} by what happens to the carried state; we call
BUT, gated BUT and BUT with input injection, which carry only an undecayed stream, the undecayed-stream models.

\paragraph{Measurements.}
Each diagnostic is a rollout to repeat 60 on the same 64 rows, compared with the unmodified rollout; full definitions are in
Appendix~\ref{app:abl-measure}. The \emph{leading Lyapunov rate} is the average log growth per repeat of an infinitesimal error
along the model's own trajectory; a negative rate means such errors shrink. \emph{Rotations} turn each token's \(x\) by an angle
\(\theta\) at one repeat, leaving its mean and norm unchanged: a small one (\(0.01\)\,rad) leaves the rows exact and gives a
halving time for the error, and a large one (\(0.6\) or \(1.0\)\,rad) eventually flips bits and tests recovery. A wrong Rule~30
row can never become the true row again, so the best a damaged rollout can do is to evolve its wrong row correctly: a repeat is
\emph{faithful} when every bit equals Rule~30 applied to the model's own previous row, and a broken branch \emph{recovers} if it
returns to five consecutive faithful repeats within 15, counted on the branches where the unperturbed branch is correct. \emph{Gate closure} skips the cell's read into \(x\) for one repeat.

\begin{table}[t]
\centering
\scriptsize
\caption{Ablations under \(R_{12}\), five seeds. \emph{Depth 60}: final-test exact-row accuracy (\%), mean \(\pm\) sample SD.
\emph{Reach}: furthest depth up to which exact-row accuracy stays at or above \(95\%\), range over seeds (evaluation stops at 60).
\emph{Rate}: leading Lyapunov rate per repeat, repeats 31--60. \emph{Halving}: repeats to halve the error in \(x\) after a \(0.01\)\,rad rotation; the cell's error is not included.
\emph{Recovery}: broken branches (\%) that recover after a \(0.6\)\,rad rotation at repeat 6. The last three columns use the 64
diagnostic rows. One gated decay BUT seed does not fit and has no reach; its row includes that seed (Appendix Table~\ref{tab:app-abl-gateddecay}).
\emph{Ours} marks the proposed model; the other rows are baselines and controls, with their sources in Section~\ref{sec:architecture}.}
\label{tab:abl-main}
\setlength{\tabcolsep}{4pt}
\renewcommand{\arraystretch}{0.9}
\begin{tabular}{@{}lrrcrrr@{}}
\toprule
Model & Params & Depth 60 & Reach & Rate & Halving & Recovery \\
\midrule
\multicolumn{7}{@{}l}{\emph{Decaying state/side state}} \\
LSTM-UT (ours) & 82,368 & \(99.67\pm0.56\) & 60 & \(-0.039\pm0.032\) & 4 & 63.1 \\
Linear memory & 57,856 & \(99.98\pm0.03\) & 60 & \(-0.082\pm0.048\) & 3 & 55.2 \\
\multicolumn{7}{@{}l}{\emph{Decaying stream}} \\
Decay BUT, \(\lambda=0.9\) & 49,472 & \(99.56\pm0.53\) & 60 & \(-0.098\pm0.007\) & 6 & 52.2 \\
Decay BUT, \(\lambda=0.5\) & 49,472 & \(99.98\pm0.04\) & 60 & \(-0.480\pm0.094\) & 1 & 81.8 \\
FPRM BUT & 49,600 & \(99.95\pm0.08\) & 60 & \(-0.999\pm0.134\) & 1 & 94.0 \\
FPRM, no re-injection & 49,600 & \(100.00\pm0.00\) & 60 & \(-0.899\pm0.140\) & 1 & 88.5 \\
\multicolumn{7}{@{}l}{\emph{Decaying gated stream}} \\
Gated decay BUT, \(\lambda=0.9\) & 57,792 & \(41.25\pm51.79\) & 24--60 & \(+0.019\pm0.113\) & 7 & 38.8 \\
\multicolumn{7}{@{}l}{\emph{Side state that never decays}} \\
No forget gate & 74,176 & \(43.17\pm50.27\) & 19--60 & \(+0.045\pm0.046\) & 5 & 80.4 \\
\multicolumn{7}{@{}l}{\emph{No decay, no side state (undecayed stream)}} \\
Gated BUT & 57,792 & \(0.00\pm0.00\) & 19--23 & \(+0.013\pm0.012\) & 95 & 0.3 \\
BUT + injection & 49,472 & \(0.00\pm0.00\) & 19--25 & \(+0.028\pm0.030\) & 32 & 10.0 \\
BUT & 49,472 & \(0.00\pm0.00\) & 18--31 & \(+0.039\pm0.055\) & 39 & 7.7 \\
\bottomrule
\end{tabular}
\end{table}

\subsection{Results}
\label{sec:abl-results}

\paragraph{BUT and CoTFormer do not extrapolate; LSTM-UT does.}
Under \(R_{12}\), BUT and CoTFormer produce no exactly correct rows at depth 60, whereas LSTM-UT reaches \(99.67\pm0.56\%\)
exact-row accuracy (Figure~\ref{fig:rule30-extrapolation}, right; \(R_6\) on the left). LSTM-UT and BUT both fit the supervised
depths, so the gap lies in extrapolation. Selecting checkpoints by in-distribution validation accuracy instead of at an
unseen depth does not change this: every BUT and gated BUT seed is at \(0\%\) exact rows at depth 60 under either rule (Appendix
Table~\ref{tab:selector}). CoTFormer is hard to fit on this task and does not extrapolate when it fits: no fitted run stays above
\(95\%\) for more than three steps past its trained depth; cutting its cache to the most recent repeat lets every run fit and extends that to four to eight steps (Appendix~\ref{app:cotformer}).

\paragraph{A decaying state extrapolates in every family but one.}
All 30 seeds of six families with a decaying state reach depth 60 (Table~\ref{tab:abl-main}), whether the decay is applied to
the stream itself (decay BUT, FPRM) or to a side state that is read back into a weight-1 stream (LSTM-UT, linear memory). The seventh, gated decay BUT, does so in two of five seeds, or two of the four that fit. All 15 seeds of the three undecayed-stream families stop between depths 18 and 31,
and wider BUTs do no better (Appendix~\ref{app:ablations}). Saturation and gates are not needed: linear memory has neither, and gated
BUT stops where BUT does. Re-anchoring to the input is neither needed nor sufficient: input injection fails, and FPRM extrapolates
with or without it. In a rollout the target depends on the current row, not on the input. Parameters do not explain it: decay BUT has
exactly BUT's 49,472. Evaluated to depth 1,000, 29 of those 30 seeds still have more than half their rows exactly right;
the exception is an LSTM-UT seed, and LSTM-UT and decay BUT with \(\lambda=0.9\) decline most (Appendix
Table~\ref{tab:depth1000}).

\paragraph{Decay is sufficient by intervention.}
Decay BUT differs from BUT only in the fixed \(\lambda\). Keeping \(90\%\) of the stream per repeat turns failure at depths 18--31
into extrapolation to depth 60 in every seed. If the block left an error untouched, the stream decay alone would shrink it by a
factor \(\lambda\) per repeat, a rate of \(\ln 0.9=-0.105\). The measured leading Lyapunov rate is \(-0.098\), so nearly all of
the contraction comes from \(\lambda\) and almost none from the learned block. At \(\lambda=0.5\) the block works against the decay: the rate is \(-0.480\) against \(\ln 0.5=-0.693\), so it undoes about a third. We trained only \(\lambda=0.9\) and \(0.5\), so we
do not know how close to 1 the decay can be.

\paragraph{The models that extrapolate contract.}
Every family that extrapolates in every seed has a negative mean rate and halves a small error in at most 6 repeats; every undecayed-stream family
has a positive mean rate and takes at least 32. Per seed the sign of the rate has three exceptions, all within \(0.004\) of zero. \citet{lu2018attractor} give the frame: a
recurrent model fitted on short horizons and then run in closed loop reproduces long-run behaviour only if perturbations transverse to
the set of valid states shrink. Fitting depths 1--12 does not require this, which is why every model fits and only some extrapolate.
A weight-1 stream with nothing else to contract it is the obstacle: BUT's linearisation is \(I+J_\Delta\), so every direction the block does not act on keeps
multiplier 1, whereas the leaky integrator has \(\lambda I+J_\Delta\) \citep{yildiz2012echo}.
The converse fails: gated decay BUT halves a small error in 7 repeats and contracts at \(-0.08\) to \(-0.10\) while its rows are exact, yet two fitted seeds stop at depth 24 (Appendix~\ref{app:abl-contraction}).

\paragraph{Every carried state has to forget.}
\citet{miller2019stable} give a sufficient condition for an LSTM to be contractive that requires its forget gate to stay below
one; removing the gate tests the case their condition excludes. Without the forget gate, \(x\) contracts as fast as in LSTM-UT, yet only two of five seeds reach depth 60. The two that do are
neutral, not contracting (late rates \(+0.001\) and \(+0.003\)), and only one of them lasts to depth 1,000. The cell keeps the error:
from the rotation to repeat 60, the gap between the perturbed and clean cells shrinks in every LSTM-UT seed and grows in every
no-forget seed (Appendix Table~\ref{tab:app-abl-cellgap}). Its high recovery does not last: in one failing seed \(92\%\) of
broken branches return to a faithful streak and \(5\%\) are still faithful at repeat 60 (Appendix~\ref{app:abl-recovery}).
\paragraph{After damage, decaying models return to faithful computation.}
We count only rows on which the unperturbed model itself stays faithful for the 15 repeats after the rotation, so a failure to
recover cannot be the model's own collapse at depth. On those rows, after a \(0.6\)\,rad rotation at repeat 6, \(52\)--\(94\%\) of
broken branches recover in the six decaying families and \(39\%\) in gated decay BUT, against at most \(10\%\) in the undecayed-stream families. Recovery is seed dependent (\(38\)--\(97\%\)
for LSTM-UT) and local: at \(1.0\)\,rad it falls to \(55\)--\(58\%\) for FPRM, \(7.5\)--\(14.5\%\) for LSTM-UT and linear memory, and
below \(1\%\) for \(\lambda=0.9\), since local contraction need not predict global return \citep{yildiz2012echo}. A state can also
be readable without being reusable, a distinction \citet{zhou2026retain} draw in looped Transformers. Right after the rotation the perturbed state still decodes to the exact correct row in at least \(99.6\%\) of
branches in every model but gated decay BUT (\(96.9\%\)), yet a later repeat goes wrong in \(56\)--\(100\%\) of them.
\paragraph{The cell is read at every step.}
After one skipped read, exact-row accuracy at repeat 60 is \(0\%\) in all 15 seeds of LSTM-UT, linear memory and the no-forget model.
The skipped repeat itself still decodes \(88\)--\(100\%\) of cells and the cell is intact, because it is written before the read; the
next, unmodified repeat is wrong. What breaks the computation is the missing term in \(x\), not lost memory. Once the read is
restored, some seeds return to applying Rule~30 to their own, now wrong, row, but lastingly in few: at least half of the branches are
still faithful at repeat 60 in 1 of 5 LSTM-UT seeds, 2 of 5 without the forget gate and 3 of 5 for linear memory. In these models, which are not trained to recall, the cell is a
fast-decaying channel used by every step, not a store of the row (Appendix~\ref{app:ablations}).

These are interventions on the architectures we trained: they show that a fixed decay is sufficient on BUT's stream, not in every architecture, and not that decay is
necessary, since a design we did not try might extrapolate another way.

\section{Delayed Cellular-Automaton Recall}
\label{sec:delayed-recall}
Delayed recall tests memory use directly: the answer is a state the model has already computed, so the question is whether it can
retrieve it. We draw conclusions only where a difference is large and holds on more than one measure.

\subsection{Task and protocol}
A model receives a binary initial row \(X^{(0)}\), executes \(H\) Rule~30 updates, and is then queried for an earlier row
\(X^{(q)}\) by its age \(a=H-q\). Evolution always completes before recall, so the model cannot answer by stopping at step
\(q\). The query is given by a learned direction and age controller,\footnote{The controller adds a two-entry direction embedding and an \((H_{\max}+1)\)-entry age embedding of width \(d\), i.e.\ \(15d\) parameters at \(H_{\max}=12\). Parameter counts in this section therefore exceed those of the same architectures on the Rule~30 task; for example, width-64 BUT has 50,432 parameters here and 49,472 there.} applied to the shared block \(\mathcal T_\theta\) as a
transient, subtractive update \(x^+=\mathcal T_\theta(x+c)-c\), where \(c=e_{\rm dir}+e_{\rm age}\).

Training supervises every evolution repeat and the recall output with cell-wise cross entropy, so the row to be recalled is one the
model was trained to compute. We train horizons 1 to \(H_{\max}\) for \(H_{\max}\in\{6,12,20\}\), with three seeds, 25,000
steps, 64-cell rows and width 64; all models share data, splits, optimiser settings and checkpoint candidates, and each run is
evaluated at the checkpoint with the best validation recall (Appendix~\ref{app:recall-protocol}). We compare BUT, CoTFormer,
LSTM-UT, gated BUT and the ungated fixed-decay models of Section~\ref{sec:ablations} (decay BUT, linear memory and FPRM BUT) and, at horizons 12 and 20, gated decay BUT. Unless stated, results
here use one recall application; two are in Appendix~\ref{app:recall-further}.

We score an answer in two ways. \emph{Cell} and \emph{exact-row} accuracy compare it with the true row \(X^{(q)}\).
\emph{Internal} accuracy compares it with the row the model itself decoded at repeat \(q\), so an error in the model's forward
computation is not counted against its recall.
\subsection{Results}
\begin{table}[t]
\centering
\footnotesize
\caption{Delayed recall after 25,000 training steps with one recall application: final-test accuracy (\%), mean \(\pm\) sample SD
over three seeds. \emph{Cell} and \emph{Exact row} compare with the true requested row. Queries for the current row are excluded and horizons are weighted equally. \emph{Reach} is the
depth-extrapolation reach of the same architecture from Table~\ref{tab:abl-main}, from separately trained models (for CoTFormer,
the furthest of any fitted run; Appendix~\ref{app:cotformer}).}
\label{tab:recall-25k}
\setlength{\tabcolsep}{2.5pt}
\renewcommand{\arraystretch}{0.9}
\scalebox{0.8}{\begin{tabular}{@{}lcrrrr@{}}
\toprule
 & & \multicolumn{2}{c}{\(H_{\max}=12\)} & \multicolumn{2}{c}{\(H_{\max}=20\)} \\
\cmidrule(lr){3-4}\cmidrule(l){5-6}
Model & Reach & Cell & Exact row & Cell & Exact row \\
\midrule
\multicolumn{6}{@{}l}{\emph{Input-dependent gates}} \\
Gated BUT & 19--23 & \(97.23\pm2.82\) & \(81.65\pm16.63\) & \(94.53\pm1.21\) & \(70.46\pm4.17\) \\
LSTM-UT (ours) & 60 & \(97.94\pm0.19\) & \(86.31\pm3.93\) & \(92.41\pm1.00\) & \(67.64\pm2.80\) \\
\multicolumn{6}{@{}l}{\emph{No gate, no decay}} \\
BUT & 18--31 & \(96.17\pm0.10\) & \(76.82\pm0.83\) & \(88.97\pm2.55\) & \(53.75\pm4.49\) \\
\multicolumn{6}{@{}l}{\emph{Fixed decay}} \\
Decay BUT, \(\lambda=0.9\) & 60 & \(96.28\pm0.47\) & \(77.67\pm8.88\) & \(89.98\pm0.54\) & \(53.71\pm8.51\) \\
Decay BUT, \(\lambda=0.5\) & 60 & \(96.47\pm0.71\) & \(80.90\pm0.88\) & \(88.13\pm0.95\) & \(44.67\pm4.02\) \\
Linear memory & 60 & \(96.20\pm0.33\) & \(79.57\pm4.89\) & \(88.43\pm3.86\) & \(49.71\pm11.79\) \\
FPRM BUT & 60 & \(95.15\pm1.78\) & \(68.81\pm9.32\) & \(86.01\pm3.46\) & \(39.99\pm6.43\) \\
\multicolumn{6}{@{}l}{\emph{Decaying gated stream}} \\
Gated decay BUT, \(\lambda=0.9\) & 24--60 & \(97.26\pm1.34\) & \(75.66\pm7.62\) & \(89.99\pm2.71\) & \(53.02\pm12.14\) \\
\multicolumn{6}{@{}l}{\emph{Full history}} \\
CoTFormer & \(\le15\) & \(89.11\pm9.51\) & \(52.13\pm31.03\) & \(86.19\pm7.15\) & \(32.16\pm28.13\) \\
\bottomrule
\end{tabular}}
\end{table}

\paragraph{Gating helps at the longest horizon; a fixed decay does not, and undoes the gate.}
At horizon 20, gated BUT and LSTM-UT recall about \(70\%\) of rows exactly; every other model is at BUT's \(54\%\) or below
(Table~\ref{tab:recall-25k}). A fixed decay never helps: on BUT's stream a decay of \(0.9\) changes nothing and stronger
contraction costs 9 to 14 points, and on gated BUT's stream \(0.9\) removes the gate's advantage. Gated models also hold older
rows: cell accuracy stays above \(95\%\) for seven to nine steps back, against four to five for BUT and decay BUT
(Figure~\ref{fig:recall-age}; gated BUT is in Appendix Figure~\ref{fig:recall-age-all}). No model recalls old rows, and at
horizons 6 and 12 the models other than CoTFormer are too close to rank with three seeds.

Together with Section~\ref{sec:ablations} this separates the two abilities. A rollout needs errors to shrink, which a fixed decay
provides on BUT's stream. Recall needs part of the state to outlast many further steps, and a fixed decay treats everything it
carries alike. Gated BUT does not extrapolate; the ungated fixed-decay models recall no better than BUT; gated decay BUT is in the leading
group on neither task; LSTM-UT is the only model here in the leading group on both. 

\begin{figure}[t]
\centering
\includegraphics[width=0.7\textwidth]{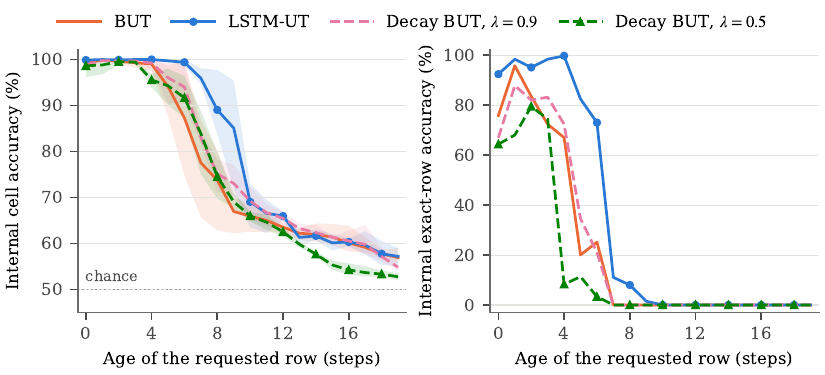}
\caption{Internal recall accuracy by the age of the requested row, after a 20-step rollout (25,000 training steps, one recall
application). Internal accuracy scores the recalled row against the row the same model decoded at the requested step, so errors of
the forward pass are not counted against recall. Left: cell accuracy; bands span the lowest and highest of three seeds. Right:
exact-row accuracy, means only. Dashed lines are the fixed-decay models; the other four models are in Appendix
Figure~\ref{fig:recall-age-all}.}
\label{fig:recall-age}
\end{figure}

\paragraph{Training difficulties and seed dependency.}
Runs are seed dependent, and this task is hard to optimise for every model: the share of rows computed correctly varies by seed in
every family, and the worst-fitted seeds are in CoTFormer and the strongly contracting models (per-seed results are in Appendix
Table~\ref{tab:recall-per-seed}). The ranking does not come from those errors: scoring recall against the row the model itself
computed changes no number by more than \(1.0\) point (Appendix Table~\ref{tab:recall-internal}). Most selected checkpoints are
at or near the final step, so the models were still improving and the ranking is one at equal training, not at convergence.
LSTM-UT's lead over BUT is 9 to 14 exact-row points (Table~\ref{tab:recall-25k}).

\subsection{Where the recalled row comes from}
\label{sec:recall-lookup}
If a model answered by looking up a stored copy of the requested row, accuracy need not depend on how long ago that row was
computed. In every model it does (Figure~\ref{fig:recall-age}): the most recent rows are recalled almost perfectly and the oldest
near chance. This is the behaviour of a trace that fades with further computation, and gates slow the fading. Interventions on the
two models with an explicit memory show where their answers come from.

\begin{figure}[t]
\centering
\includegraphics[width=0.7\textwidth]{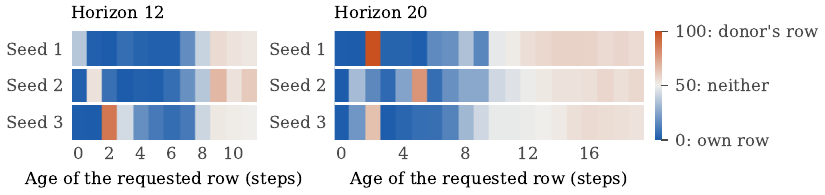}
\caption{Swapping LSTM-UT's cell between two inputs just before the recall application (25,000 training steps, one recall
application; one row per trained model). Colour is the donor share: of the cells where the
input's own target and the donor's target differ, the percentage on which the output equals the donor's. 0 means the foreign cell is
ignored and the input's own row is returned, 50 that the output carries no information, and 100 that the output is the donor's row.}
\label{fig:cell-swap}
\end{figure}

\paragraph{CoTFormer looks the row up at horizon 6 and not at horizon 12.}
CoTFormer stores every earlier state and is last in exact rows at both horizons. Much of this is fitting, not recall: at horizon 20
no seed computes its final row correctly. Its cache holds the completed row \(q\) in the block written at repeat \(q+1\), and we
gave that block another input's values, or hid it. At horizon 6, where every seed recalls almost perfectly, two of three seeds look
the row up: for most queries the swap takes recall to chance, and swapping any other block does nothing. At horizon 12 the lookup
is gone: on the checkpoints that recall well, swapping or hiding the block changes recall by at most \(2.2\) points. At horizon 20
the test is not informative, since two seeds compute almost no rows past depth 8. A stored history can be read as addressable
memory, but we found that solution only where the task was easiest (Appendix~\ref{app:cotformer}).
\paragraph{LSTM-UT reads some rows from its cell.}
\label{sec:recall-cell}
We intervened on the cell just before the recall application, on the six selected LSTM-UT checkpoints of the 25,000-step runs
(three seeds at horizons 12 and 20). Setting the cell to zero takes exact rows from \(86.3\%\) to \(58.6\%\) and from
\(67.6\%\) to \(38.7\%\), while the current-row query, which needs no memory, stays at \(99\%\) of cells: the cell matters
for recall and zeroing it does not break the step. To test whether it holds the rows themselves, we swapped it between two
unrelated inputs, A and B, and asked A for an earlier row while it held B's cell (Figure~\ref{fig:cell-swap}). On the cells where
the two correct answers differ, the \emph{donor share} is the percentage of output bits that equal B's: \(0\%\) means A's own
row came back, \(100\%\) B's. Every checkpoint has ages at which the output follows the donor; in one horizon-20 model the row
from two steps back is the donor's row exactly for \(99.6\%\) of test inputs. At the other ages the donor share stays near zero,
so those rows are held in the stream. LSTM-UT therefore divides memory between its stream and its cell, and which ages go to the
cell differs between trained models (Appendix~\ref{app:cell-interventions}).

\section{Language Modelling}
\label{sec:lm}
The cellular-automaton tasks isolate state tracking; we also check that LSTM-UT trains competitively as a language model, and at
what cost. We train GPT-2-style models on OpenWebText2 following the CoTFormer protocol \citep{mohtashami2025cotformer}: a 2-layer
prelude, a 9-layer core looped 5, 8 or 10 times, and a 1-layer coda, at width 384, with one seed per configuration
(Appendix~\ref{app:lm}). LSTM-UT reaches lower validation perplexity than BUT at every repeat count, by 1.2 to 1.3 points, and
slightly lower than CoTFormer at equal width (29.90, 29.65 and 29.67 against 30.22, 29.84 and 29.82). With parameters matched at
five repeats CoTFormer is better (29.31 against 29.90), and with a layer normalisation after every repeat LN-CoTFormer is lower at 8
and 10 repeats. CoTFormer takes \(22.5\%\), \(60.4\%\) and \(90.7\%\) more time per training step than LSTM-UT at 5, 8 and 10
repeats, because each repeat attends over every earlier repeat; LSTM-UT costs \(10.8\%\) to \(13.6\%\) more than BUT. With one
seed, differences of a few tenths are indicative only. The evidence is that a bounded gated memory is competitive with an expanding
cache at a fraction of its training cost, not that it improves perplexity over CoTFormer.

\section{Discussion and Limitations}
What the LSTM cell holds is resolved only in part. In the recall models, swapping the cell between inputs shows that it
carries the content of some earlier rows, at ages that differ between trained models, while other rows are carried by the stream
(Section~\ref{sec:recall-lookup}). We do not know how a row is encoded in the cell, why training assigns a given age to the cell
in one seed and to the stream in another, or whether the same holds on other tasks. In the
extrapolation models, which are never asked to recall, the cell behaves as a channel used by every step
(Section~\ref{sec:ablations}).
We claim that stability and recall have separate requirements, not that one is traded against the other. Each recall setting has
three seeds. At horizon 20 exact-row recall falls in the order of decay strength (BUT and \(\lambda=0.9\), then \(\lambda=0.5\), then
FPRM BUT), but neighbouring models overlap in their spread, the order among BUT and the two decay BUTs reverses at horizon 12, and FPRM BUT differs from BUT in more
than its decay. On BUT's stream a decay of \(0.9\) costs nothing at any horizon. A fixed decay on gated BUT's stream, the simplest combination without a cell, 
extrapolates in two of five seeds and recalls at BUT's level, so the two ingredients alone are not enough.
In LSTM-UT the decay is itself gated and acts on a state the block does not see.

The evidence is restricted to deterministic binary cellular automata, with three to five model seeds per setting and a shared data
seed. The language-modelling runs use one seed per configuration and test training quality and cost, not memory or extrapolation.
Rule~30 does not reproduce partial observability, stochastic dynamics or open-ended outputs, so the results make a contracting,
gated state a candidate for those settings, not an established advantage.

\section{Conclusion}

A looped transformer used as a rollout must stay correct past its trained horizon and keep earlier states retrievable. On Rule~30 these depend
on different properties of the state the loop carries. Rollouts stay exact when that state contracts: a fixed decay on BUT's stream is
enough, though not on a gated stream, and no model without a decaying state extrapolates reliably. Recall is not helped by any fixed decay and is helped by a gate computed
from the input, unless the gated stream also decays at a fixed rate. Contraction for stability and input-dependent retention for memory is the combination the LSTM reached over time with its forget gate;
LSTM-UT applies it over depth and is the only model we tested in the leading group on both tasks. Storing every earlier state
substituted for neither property: the stored copy was looked up at the shortest horizon and not at the next.

More broadly, treating the loop itself as the rollout turns questions about a looped transformer into questions about a dynamical
system: whether its errors shrink, what it can still retrieve, and where that information is held. On Rule~30 these have different
answers for stability and for memory, and each came from intervening on the carried state, not from comparing architectures.

\section*{Reproducibility statement}
The cellular-automaton data are generated, not collected: every training, validation and test row follows from a recorded seed, and
Appendix~A gives the seeds, the split sizes and the check that the splits share no rows. Section~\ref{sec:rule30} and Appendix~A
give the architecture, optimiser settings and checkpoint-selection rule for depth extrapolation, Section~\ref{sec:delayed-recall} and
Appendix~\ref{app:recall-protocol} those for delayed recall, and Appendix~\ref{app:lm} the language-modelling setup, which follows
the public CoTFormer protocol on OpenWebText2. Every model variant is defined by its update equation in
Section~\ref{sec:architecture} and Appendix Table~\ref{tab:models}, and every diagnostic in
Appendices~\ref{app:abl-measure} and~\ref{app:cell-interventions}. Results are reported over three to five model-initialisation
seeds, with the spread or the per-seed values given. The code, the run configurations and the scripts that launch the experiments are available at
\url{https://github.com/arskvnc22/lstm-ut}.

\section*{AI use statement}
We used generative AI tools (large language model writing and coding assistants) in preparing this paper. For writing, they helped
restructure and edit the manuscript, including drafting revised versions of the abstract, introduction, section openings, and conclusion
from our results, and checking the text for internal consistency. For code and analysis, they wrote the plotting and table scripts and
assisted with verification: recomputing reported statistics from logged run metrics, regenerating the Rule~30 data splits to test for
overlap between training, validation, and test data, and the boundary-cell analysis of early-depth behaviour in
Appendix~\ref{app:early-depth}. AI coding assistants also helped implement parts of the training and evaluation code. All reported
numbers come from our own training runs; AI-generated code and analyses were verified by comparing their outputs against the logged
metrics and the reported tables. We have reviewed all AI-assisted work and take responsibility for the final content of this work,
including text, claims, and artifacts produced with the aid of generative AI.

\bibliography{references}
\bibliographystyle{iclr2027/iclr2027_conference}

\appendix
\raggedbottom
\section{Additional experimental details}

\subsection{Training and checkpoint provenance}
The Rule~30 initial rows are independent Bernoulli$(0.5)$ samples with periodic boundaries. Training rows are materialised once from a fixed data seed and shuffled with a fixed shuffle seed; target rows are recomputed exactly for the selected depth. Each validation and final-test split contains 4,096 independently generated rows. Model-initialisation seeds vary within a comparison, while data, shuffle, validation, and test seeds are held fixed. Depth pairs are cycled round-robin by optimizer step rather than mixed within a batch. Runs trained on depths 1--4 used learning rate $10^{-3}$, weight decay $0.15$, and gradient clipping at $0.9$; the deeper regimes changed weight decay to $0.1$. All runs used zero dropout. The unconstrained extrapolation checkpoints used in Tables~\ref{tab:rule30-r6-r8-extrapolation}--\ref{tab:rule30-r12-extrapolation} maximise validation cell accuracy at depth 9 for $R_6$ and at depth 15 for $R_{12}$.
The final-test set did not select checkpoints. All depth-extrapolation results are reported over five model-initialisation seeds.

\begin{table}[htbp]
\centering
\scriptsize
\setlength{\tabcolsep}{3pt}
\caption{$R_6$ Rule 30 depth extrapolation using the unconstrained extrapolation-selected checkpoint. Each result reports cell accuracy and exact-row accuracy, in percent, as mean $\pm$ sample standard deviation over five model-initialisation seeds. Depth 6 is the supervised boundary, depth 9 is used for checkpoint selection, and depths 12 and 20 test further extrapolation.}
\label{tab:rule30-r6-r8-extrapolation}
\begin{tabular}{@{}lcccrrrr@{}}
\toprule
Architecture & $d$ & Seeds & Metric & Depth 6 & Depth 9 & Depth 12 & Depth 20 \\
\midrule
\multirow{2}{*}{LSTM-UT} & \multirow{2}{*}{32} & \multirow{2}{*}{5}
& Cell & $99.998\pm0.004$ & $99.990\pm0.015$ & $99.975\pm0.041$ & $99.790\pm0.361$ \\[2pt]
& & & Exact & $99.922\pm0.124$ & $99.741\pm0.422$ & $99.370\pm1.043$ & $96.514\pm6.249$ \\
\cmidrule(lr){1-8}
\multirow{2}{*}{LSTM-UT} & \multirow{2}{*}{64} & \multirow{2}{*}{5}
& Cell & $100.000\pm0.000$ & $99.999\pm0.001$ & $99.977\pm0.018$ & $98.774\pm1.042$ \\[2pt]
& & & Exact & $100.000\pm0.000$ & $99.951\pm0.057$ & $99.131\pm0.647$ & $74.863\pm19.437$ \\
\cmidrule(lr){1-8}
\multirow{2}{*}{BUT} & \multirow{2}{*}{64} & \multirow{2}{*}{5}
& Cell & $99.999\pm0.002$ & $99.980\pm0.025$ & $98.546\pm2.264$ & $65.705\pm19.859$ \\[2pt]
& & & Exact & $99.956\pm0.074$ & $99.062\pm1.181$ & $67.056\pm41.173$ & $0.449\pm0.655$ \\
\cmidrule(lr){1-8}
\multirow{2}{*}{CoTFormer} & \multirow{2}{*}{64} & \multirow{2}{*}{5}
& Cell & $88.254\pm20.780$ & $64.430\pm20.683$ & $50.002\pm0.057$ & $50.001\pm0.078$ \\[2pt]
& & & Exact & $51.787\pm50.042$ & $3.735\pm8.353$ & $0.000\pm0.000$ & $0.000\pm0.000$ \\
\bottomrule
\end{tabular}
\end{table}

\begin{table*}[htbp]
\centering
\scriptsize
\setlength{\tabcolsep}{2.5pt}
\caption{$R_{12}$ Rule 30 depth extrapolation using the unconstrained extrapolation-selected checkpoint. Results are percentages reported as mean $\pm$ sample standard deviation over five model-initialisation seeds. Depth 12 is the supervised boundary, depth 15 selects the checkpoint on validation data, and depths 40 and 60 are beyond both training and checkpoint selection. The width-80 BUT and width-48 LSTM-UT rows approximately match the parameter count of width-64 LSTM-UT and BUT, respectively.}
\label{tab:rule30-r12-extrapolation}
\begin{tabular}{@{}lrrrcrrrr@{}}
\toprule
Architecture & \(d\) & Parameters & Seeds & Metric & Depth 12 & Depth 20 & Depth 40 & Depth 60 \\
\midrule
\multirow{2}{*}{LSTM-UT} & \multirow{2}{*}{64} & \multirow{2}{*}{82,368} & \multirow{2}{*}{5}
& Cell & \(100.000\pm0.000\) & \(100.000\pm0.000\) & \(99.971\pm0.056\) & \(99.916\pm0.152\) \\[2pt]
& & & & Exact & \(100.000\pm0.000\) & \(99.995\pm0.011\) & \(99.868\pm0.230\) & \(99.673\pm0.557\) \\
\cmidrule(lr){1-9}
\multirow{2}{*}{BUT} & \multirow{2}{*}{64} & \multirow{2}{*}{49,472} & \multirow{2}{*}{5}
& Cell & \(100.000\pm0.001\) & \(99.783\pm0.383\) & \(55.354\pm10.616\) & \(50.053\pm0.098\) \\[2pt]
& & & & Exact & \(99.995\pm0.011\) & \(93.149\pm11.626\) & \(0.000\pm0.000\) & \(0.000\pm0.000\) \\
\midrule
\multirow{2}{*}{BUT} & \multirow{2}{*}{80} & \multirow{2}{*}{77,200} & \multirow{2}{*}{5}
& Cell & \(99.933\pm0.140\) & \(99.658\pm0.556\) & \(57.979\pm17.762\) & \(50.010\pm0.037\) \\[2pt]
& & & & Exact & \(99.185\pm1.690\) & \(92.959\pm8.831\) & \(2.812\pm6.289\) & \(0.000\pm0.000\) \\
\cmidrule(lr){1-9}
\multirow{2}{*}{LSTM-UT} & \multirow{2}{*}{48} & \multirow{2}{*}{46,416} & \multirow{2}{*}{5}
& Cell & \(95.291\pm5.414\) & \(91.484\pm9.549\) & \(80.624\pm18.941\) & \(72.009\pm22.109\) \\[2pt]
& & & & Exact & \(54.644\pm50.855\) & \(53.594\pm49.864\) & \(44.316\pm44.147\) & \(38.506\pm44.505\) \\
\bottomrule
\end{tabular}
\end{table*}

We also ran an intermediate \(R_8\) regime (width 64, five seeds). It showed the same ordering: at depth 60, LSTM-UT reached
\(73.9\pm42.8\%\) exact-row accuracy against \(0\%\) for BUT. We omit its table for space.

For delayed recall, all models use AdamW with learning rate $10^{-3}$, weight decay $0.15$, gradient-norm clipping at $0.9$, batch size 128, and no learning-rate 
schedule. Every training batch contributes evolution and query losses; intermediate evolution targets are exact Rule~30 rows, but clean states are
never fed into the recurrence. The main sweep covers maximum horizons 6, 12, and 20 with three model-initialisation seeds. The source data
seed is shared across seeds and architectures. The primary checkpoint maximises validation recall cell accuracy under the pair-balanced macro, 
with worst-query cell accuracy, exact-row accuracy, forward accuracy, and loss as deterministic tie-breakers. Test data do not select checkpoints.

\subsection{Evaluation diagnostics}
Cell accuracy is the fraction of correctly predicted cells; exact-row accuracy requires all 64 cells to match. For delayed recall, queries are averaged within each horizon and the horizon means are weighted equally, excluding the no-op query $q=H$. Internal accuracy compares the recalled row with the model's own decoded row at requested repeat $q$, rather than the exact automaton row. It must be interpreted alongside ground-truth accuracy because a collapsed recurrent trajectory can yield high self-agreement without correct recall.

Changed-cell internal accuracy restricts that comparison to positions at which the model's requested-repeat and current-repeat decoded rows differ. On these cells, matching the current binary value necessarily fails to match the requested one. This makes the diagnostic more sensitive to selection than raw internal accuracy, though it can be unstable when few cells differ. The unique-best statistic compares the recalled row with the true row at every repeat of the trajectory and counts a success only when the requested row is strictly the closest in cell agreement. Ties are not successes. Unlike the two internal measures, it is scored against the true rows. The cosine-best statistic is its internal counterpart: it compares the logits of the answer with the model's own logits at every rollout repeat, by cosine similarity, and counts a success only when the requested repeat is strictly the closest. These diagnostics help separate forward-computation failure from failure to select an earlier computed state, but they do not reveal a unique internal mechanism.

\subsection{Width, parameter count and long-horizon stress tests}
\label{app:width}
Parameter count does not explain the gap. Width-80 BUT (77,200 parameters, within \(6.3\%\) of width-64 LSTM-UT) still fell to \(50.01\%\)
cell and \(0\%\) exact-row accuracy at depth 60. Nor does extrapolation simply shrink with width. Under \(R_6\), the 20,704-parameter width-32
LSTM-UT, with fewer than half the parameters of width-64 BUT, extrapolated further than width-64 LSTM-UT: \(96.5\pm6.2\%\) against
\(74.9\pm19.4\%\) exact-row accuracy at depth 20. Under \(R_{12}\), the 46,416-parameter width-48
LSTM-UT can also extrapolate: one seed was exactly correct on every final-test row at every depth from 2 through 60, although at depth 1 it
computed the update with non-wrapping row edges (Appendix~\ref{app:early-depth}). It does so unreliably, with two seeds collapsing to chance and two intermediate. Width therefore affects how
reliably training finds the extrapolating solution, but not monotonically, and that solution does not require the width-64 parameter count.

\paragraph{Long-horizon stress tests.}
We evaluated all five \(R_{12}\) LSTM-UT seeds through depth 1,000 on three further data seeds of 4,096 rows at the training density
\(p=0.5\), and on four further densities \(p\in\{0.1,0.25,0.75,0.9\}\). At \(p=0.5\), four of the five seeds still had more than half
their rows exactly right at depth 1,000 (\(94.6\), \(93.7\), \(68.2\) and \(99.98\%\)); the fifth stayed above \(97\%\) to depth 200
and fell below \(50\%\) at depth 491; it is the one LSTM-UT seed whose late leading rate is not negative
(Appendix Table~\ref{tab:app-abl-cellgap}). By comparison, BUT and gated BUT fell below \(50\%\) by depths 21--36 and 23--29, and to \(0\%\)
by depths 23--40, and were not evaluated further. The best seed kept above \(99.95\%\) exact-row accuracy at every depth and
density; at depth 1,000 only 10 of its 61,440 rows were wrong. An \(R_8\) checkpoint likewise retained \(99.95\%\) at depth 200. At
these depths, cell accuracy is uninformative: a wrong row has \(46.6\)--\(51.7\%\) of its cells right, chance level, because a single
Rule~30 error spreads one cell per step in each direction and covers the 64-cell ring within about 32 steps. We therefore report
exact-row accuracy.

Table~\ref{tab:depth1000} gives the same evaluation at \(p=0.5\) for all nine families evaluated to depth 1,000. Every seed
of linear memory, FPRM BUT with and without re-injection, and decay BUT with \(\lambda=0.5\) keeps at least \(98.3\%\) of rows
exact at depth 1,000. Decay BUT with \(\lambda=0.9\) and the linear-read variant decline more, to \(54.8\)--\(99.8\%\) and
\(82.8\)--\(99.9\%\). Without the forget gate one seed keeps \(99.8\%\) and the other four fall below \(50\%\) by depths 24--134
and are at \(0\%\) at depth 1,000. Gated decay BUT is as seed dependent here as at depth 60: one seed has every row exact at depth
1,000, one declines slowly to \(23.1\%\), falling below \(50\%\) at depth 481, and the other three fall below \(50\%\) by depths 21--36
and are at \(0\%\).

\begin{table}[htbp]
\centering
\small
\caption{Exact-row accuracy (\%) far beyond the trained depth, for the \(R_{12}\) checkpoints of Table~\ref{tab:abl-main} at the
training density \(p=0.5\), pooled over three further data seeds of 4,096 rows (12,288 rows per model seed). \emph{Depth 200} and
\emph{Depth 500}: range over the five model seeds. \emph{Depth 1,000}: each seed. \emph{Seeds}: seeds with at least \(50\%\) of
rows exact at depth 1,000. BUT, gated BUT and BUT + injection are at \(0\%\) by depth 60 and were not evaluated to depth 1,000.}
\label{tab:depth1000}
\setlength{\tabcolsep}{4pt}
\begin{tabular}{@{}lccrrrrrc@{}}
\toprule
 & & & \multicolumn{5}{c}{Depth 1,000, seed} & \\
\cmidrule(lr){4-8}
Model & Depth 200 & Depth 500 & 0 & 1 & 2 & 3 & 4 & Seeds \\
\midrule
LSTM-UT & 93.4--100.0 & 48.2--100.0 & 94.59 & 93.72 & 68.20 & 4.74 & 99.98 & 4/5 \\
Linear read & 96.6--100.0 & 91.1--100.0 & 99.86 & 99.70 & 97.62 & 89.75 & 82.76 & 5/5 \\
Linear memory & 99.9--100.0 & 99.5--100.0 & 99.02 & 100.00 & 99.99 & 100.00 & 99.64 & 5/5 \\
Decay BUT, \(\lambda=0.9\) & 90.1--100.0 & 74.9--99.9 & 83.15 & 96.21 & 54.76 & 93.72 & 99.77 & 5/5 \\
Decay BUT, \(\lambda=0.5\) & 99.7--100.0 & 99.5--100.0 & 100.00 & 100.00 & 99.85 & 98.98 & 99.93 & 5/5 \\
FPRM BUT & 99.6--100.0 & 99.2--100.0 & 99.99 & 98.28 & 99.80 & 99.96 & 100.00 & 5/5 \\
FPRM, no re-injection & 100.0 & 100.0 & 100.00 & 100.00 & 99.98 & 100.00 & 99.98 & 5/5 \\
No forget gate & 0.0--100.0 & 0.0--99.9 & 0.00 & 99.81 & 0.00 & 0.00 & 0.00 & 1/5 \\
Gated decay BUT, \(\lambda=0.9\) & 0.0--100.0 & 0.0--100.0 & 0.00 & 100.00 & 0.00 & 0.00 & 23.11 & 1/5 \\
\bottomrule
\end{tabular}
\end{table}

\subsection{Early-depth non-monotonicity in Rule~30 extrapolation}
\label{app:early-depth}
In Figure~\ref{fig:rule30-extrapolation} (right), the width-48 LSTM-UT and width-80 BUT means are lower at depths 1--2 than at later
supervised depths. Each depth is a separate forward pass of exactly \(d\) block applications from the initial row, decoded only at the
end; decoded rows are never fed back into the recurrence, so a depth-1 output error need not appear at depth 2.

The evaluator records accuracy at the two boundary cells (positions 0 and 63, whose periodic neighbours wrap around the row) separately
from the 62 interior cells. Table~\ref{tab:early-depth-boundary} shows that every depth-1 error in the affected runs lies at a boundary
cell: interior accuracy is exactly \(100\%\) over all 4,096 final-test rows. The errors match Rule~30 computed with fixed zero edges
instead of periodic ones. Applied to the same final-test rows, that rule gives \(23.36\%\) exact-row accuracy, \(61.68\%\) boundary
accuracy, and 0.766 errors per row; seed 0 of width-48 LSTM-UT gives \(23.3\%\), \(61.57\%\), and 0.769. Its logged test example is wrong
only at cell 63, as the zero-edge rule predicts for that row. Seeds 1 and 3 match equally closely, while the remaining affected seeds lie
between the zero-edge rule and the correct periodic update. From depth 2 onward, width-48 seed 0 is exact at both boundary and interior
cells through depth 60.

\begin{table}[htbp]
\centering
\small
\caption{Depth-1 behaviour of the \(R_{12}\) runs in Figure~\ref{fig:rule30-extrapolation} that are not exact at depth 1
(final test, unconstrained extrapolation checkpoint, 4,096 rows). Boundary accuracy averages cells 0 and 63; interior accuracy
averages cells 1--62. All width-64 LSTM-UT seeds, width-80 BUT seeds 0--2, and width-64 BUT seeds 1--4 are exact at depth 1.
The final row applies Rule~30 with zero edges to the same test rows.}
\label{tab:early-depth-boundary}
\begin{tabular}{@{}lrrrrr@{}}
\toprule
Run & Depth-1 exact & Boundary & Interior & Errors/row & Depth-2 exact \\
\midrule
LSTM-UT \(d=48\), seed 0 & \(23.3\) & \(61.57\) & \(100.00\) & 0.769 & \(100.0\) \\
LSTM-UT \(d=48\), seed 1 & \(26.0\) & \(62.77\) & \(100.00\) & 0.745 & \(42.2\) \\
LSTM-UT \(d=48\), seed 2 & \(37.7\) & \(63.21\) & \(100.00\) & 0.736 & \(98.6\) \\
LSTM-UT \(d=48\), seed 3 & \(25.4\) & \(61.88\) & \(100.00\) & 0.762 & \(17.1\) \\
LSTM-UT \(d=48\), seed 4 & \(41.7\) & \(62.84\) & \(100.00\) & 0.743 & \(36.5\) \\
BUT \(d=80\), seed 3 & \(50.0\) & \(62.39\) & \(100.00\) & 0.752 & \(72.9\) \\
BUT \(d=80\), seed 4 & \(37.5\) & \(68.42\) & \(100.00\) & 0.632 & \(73.9\) \\
BUT \(d=64\), seed 0 & \(74.8\) & \(87.40\) & \(100.00\) & 0.252 & \(100.0\) \\
\midrule
Zero-edge Rule~30 & \(23.36\) & \(61.68\) & \(100.00\) & 0.766 & -- \\
\bottomrule
\end{tabular}
\end{table}

The behaviour is not an artefact of checkpoint selection: the in-distribution-selected checkpoints of the same runs show the same depth-1
deficit. It is also present during training. In width-48 seed 0, the depth-1 training loss at step 23,000 is \(0.0166\), against
\(3.2\times10^{-4}\) at depth 2, and validation exact-row accuracy at depth 1 stays between roughly \(24\%\) and \(51\%\) throughout
training while depths 2 and 12 approach \(100\%\). The \(R_6\) width-48 LSTM-UT runs, which are not plotted, show the same pattern
(seed 0: \(25.2\%\) depth-1 exact-row accuracy, \(62.39\%\) boundary accuracy, \(100\%\) interior accuracy).

We also checked that these curves are not affected by data leakage. Regenerating the training, validation, and final-test splits from
their recorded seeds shows no shared rows among them, for both the 100,000-row \(R_6\) and 500,000-row \(R_{12}\) training sets. No
validation or test row occurs on any training row's Rule~30 trajectory through depth 60. The regenerated test and validation rows match
the examples printed in the run logs. Checkpoint selection uses only the validation split, and targets at every depth are computed
with the same periodic Rule~30 update in training and evaluation. These checks rule out leakage and target mismatch as explanations. They
do not explain why these models omit the wrap-around at the first application but include it from the second onward.

\section{CoTFormer: fitting, extrapolation and cache use}
\label{app:cotformer}
This appendix collects the CoTFormer evidence summarised in Sections~\ref{sec:rule30} and~\ref{sec:delayed-recall}.

\subsection{Fitting and reach on Rule~30}
Table~\ref{tab:cotf-fit} lists every width-64 CoTFormer run we completed on Rule~30. A run \emph{fits} if it has at least \(95\%\)
final-test exact rows at its deepest trained depth. With the full cache, at most half of the runs fit at every training range except
depths 1--6 and 1--8, and no fitted run stays at or above \(95\%\) for more than three steps beyond its trained depth, or above
\(50\%\) for more than four. Unseen-depth selection gives the same or a shorter reach for every fitted full-cache run. Restricting
attention to the most recent repeat, which removes the history, is the only setting in which every run fits and extrapolates.

\begin{table}[htbp]
\centering
\small
\caption{CoTFormer on Rule~30 at width 64, under in-distribution checkpoint selection. \emph{Fit}: runs with at least \(95\%\)
final-test exact rows at the deepest trained depth. \emph{Reach} and \emph{Half}: range over fitted runs of the furthest depth up to
which exact-row accuracy stays at or above \(95\%\) and \(50\%\). Full-cache rows count every completed run, including early pilots
and two replications with a different attention implementation. Window rows restrict the cache to the most recent \(w\) repeats in
training and evaluation, and were evaluated to depth 20. They also count three earlier runs that repeat a seed of the three-seed
sweep in Table~\ref{tab:cotf-cache-policies-short} (one at \(w=1\), two at \(w=4\)).}
\label{tab:cotf-fit}
\begin{tabular}{@{}llrrcc@{}}
\toprule
Cache & Trained depths & Runs & Fit & Reach & Half \\
\midrule
Full & 1--3 & 2 & 1 & 3 & 3 \\
Full & 1--4 & 47 & 15 & 4--5 & 4--5 \\
Full & 1--6 & 9 & 5 & 6--8 & 7--9 \\
Full & 1--8 & 3 & 3 & 8--9 & 10 \\
Full & 1--12 & 10 & 3 & 12--15 & 14--16 \\                 
\midrule
\(w=1\) & 1--12 & 4 & 4 & 16--20 & 20 \\
\(w=2\) to \(8\) & 1--12 & 23 & 14 & 12--16 & 14--17 \\
\bottomrule
\end{tabular}
\end{table}

\paragraph{The two selection rules under \(R_{12}\).}
Of the five CoTFormer seeds in Figure~\ref{fig:rule30-extrapolation}, one fits. Selected in distribution it has \(98.8\), \(97.9\),
\(93.9\), \(67.7\) and \(11.7\%\) exact rows at depths 12 to 16; selected at the unseen depth it has \(97.3\), \(96.5\), \(93.8\),
\(84.8\) and \(50.7\%\); both are at \(0\%\) by depth 18. A second seed reaches \(92.4\%\) at depth 12 under both rules and three
never train. An earlier run of the same protocol, for which both rules pick the same checkpoint, has \(99.8\), \(99.7\), \(98.6\),
\(96.0\) and \(84.1\%\) at depths 12 to 16 and \(0\%\) at depth 18; it is the furthest-reaching full-cache run we have. Under
unseen-depth selection 2 of the 10 full-cache \(R_{12}\) runs fit, against 3 in distribution. Table~\ref{tab:selector} compares the
two rules for the other models.

\begin{table}[htbp]
\centering
\small
\caption{In-distribution against unseen-depth checkpoint selection under \(R_{12}\), five seeds per model. \emph{Same}: seeds for
which both rules select the same checkpoint. \emph{Reach}: range over seeds, as in Table~\ref{tab:abl-main}; for CoTFormer, the one
seed that fits. Exact-row accuracy at depth 60 is \(98.68\)--\(100.00\%\) for LSTM-UT and \(0\%\) for the other three under both
rules.}
\label{tab:selector}
\begin{tabular}{@{}lccc@{}}
\toprule
Model & Same & Reach, in-distribution & Reach, unseen depth \\
\midrule
LSTM-UT & 5 of 5 & 60 & 60 \\
BUT & 4 of 5 & 18--24 & 18--31 \\
Gated BUT & 2 of 5 & 18--23 & 19--23 \\
CoTFormer & 1 of 5 & 13 & 13 \\
\bottomrule
\end{tabular}
\end{table}

\subsection{Cache windows and state--cache interventions}
CoTFormer can attend to every earlier step, yet it extrapolated least far; we ask whether that history helps or hurts.
Under the \(R_{12}\) protocol, CoTFormer showed a gradual failure regime suitable for intervention: its strongest run fell from \(99.93\%\) cell accuracy at repeat 15 to \(50.49\%\) at repeat 20. Restricting a frozen full-cache checkpoint to recent cached repeats did not rescue it. Training with bounded cache visibility also did not produce a monotonic cache-size benefit: at depth 20, full-cache accuracy was \(50.08\pm0.01\%\), every recent-2 through recent-8 policy was near chance, while recent-1, functionally equivalent to BUT, retained \(99.39\pm0.55\%\) (Table~\ref{tab:cotf-cache-policies-short}). 

\begin{figure}[!t]
\centering
\includegraphics[width=0.75\textwidth]{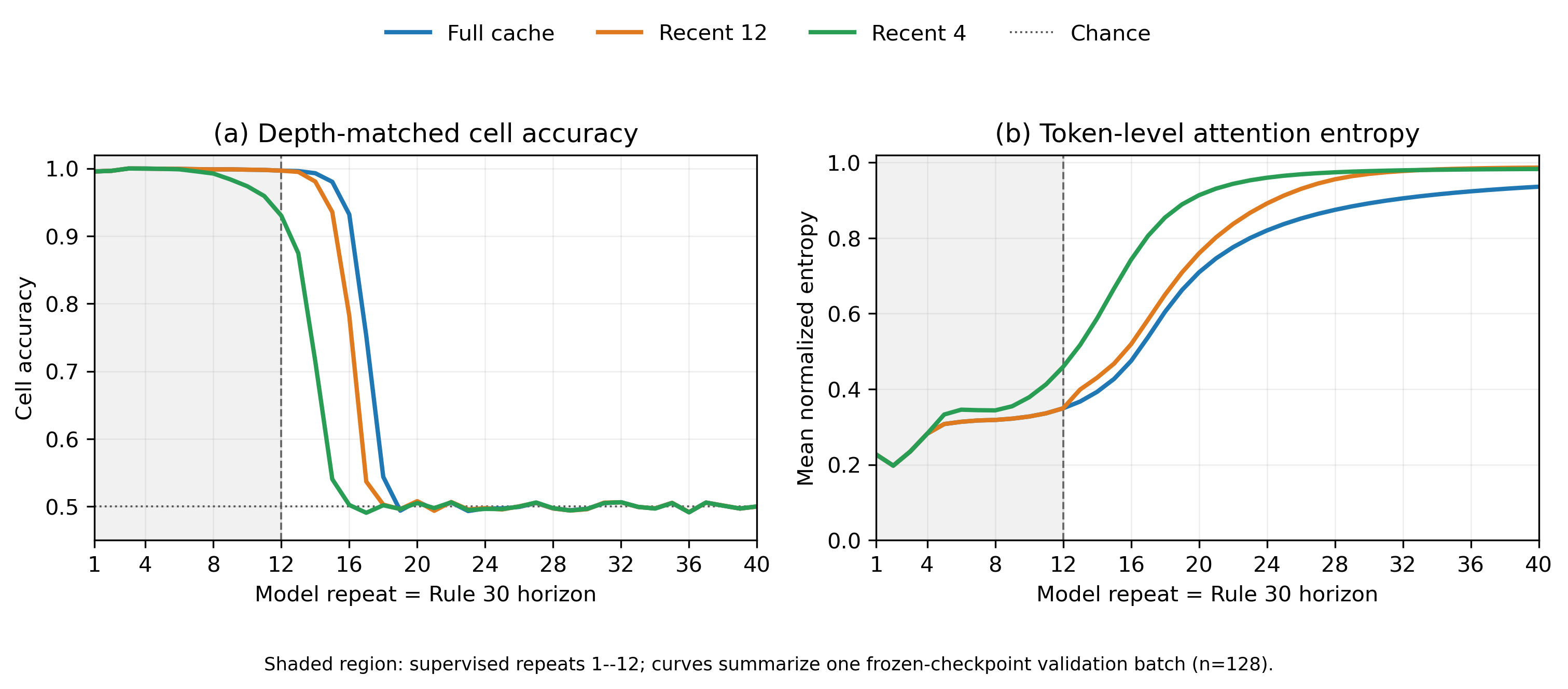}
\caption{Representative CoTFormer accuracy and normalised attention entropy under inference-time cache masking. The frozen checkpoint was not retrained for the restricted windows, so this is a diagnostic intervention rather than a trained-policy comparison.}
\label{fig:cotf-entropy-appendix}
\end{figure}

The $R_{12}$ CoTFormer runs were trained for 24,000 optimizer steps on 500,000 initial rows, supervising depths 1 through 12. In one manual-attention checkpoint, the newest three cache blocks received $75.04\%$ of attention mass at repeat 16, the earliest three received $16.24\%$, and the ten intermediate blocks only $8.72\%$. From repeat 12 to 40, normalised token entropy rose from $0.349$ to $0.936$ while mean maximum attention probability fell from $0.432$ to $0.0027$. These correlations do not establish whether attention flattening caused the representational failure or followed it. Masking the frozen checkpoint to recent-12 or recent-4 blocks failed to rescue extrapolation (Figure~\ref{fig:cotf-entropy-appendix}); this inference-time distribution shift is why separately trained cache-window models were also tested
(Table~\ref{tab:cotf-cache-policies-short}).

\begin{table}[htbp]
\centering
\small
\setlength{\tabcolsep}{4pt}
\caption{CoTFormer cell accuracy (\%) under cache policies active throughout $R_{12}$ training, mean $\pm$ sample SD. Checkpoints were selected by unseen-depth validation accuracy. Window rows are a three-seed sweep; the full-cache row is the two full-cache runs that use the same attention implementation as the window runs. Table~\ref{tab:cotf-fit} counts these and all other completed runs.}
\label{tab:cotf-cache-policies-short}
\begin{tabular}{@{}lcrrr@{}}
\toprule
Training cache & Seeds & Depth 12 (trained) & Depth 17 & Depth 20 \\
\midrule
Full & 2 & $99.74\pm0.09$ & $65.63\pm13.41$ & $50.08\pm0.01$ \\
Recent 1 & 3 & $99.99\pm0.02$ & $99.88\pm0.14$ & $99.39\pm0.55$ \\
Recent 2 & 3 & $99.86\pm0.10$ & $62.36\pm10.09$ & $49.97\pm0.10$ \\
Recent 3 & 3 & $83.25\pm28.72$ & $62.37\pm18.85$ & $49.95\pm0.14$ \\
Recent 4 & 3 & $99.73\pm0.40$ & $75.94\pm15.64$ & $50.08\pm{<}0.01$ \\
Recent 5 & 3 & $99.22\pm1.21$ & $70.33\pm25.49$ & $49.99\pm0.12$ \\
Recent 6 & 3 & $83.15\pm28.75$ & $56.29\pm10.25$ & $50.04\pm0.06$ \\
Recent 7 & 3 & $95.52\pm6.99$ & $64.74\pm17.05$ & $49.99\pm0.08$ \\
Recent 8 & 3 & $83.23\pm28.82$ & $60.55\pm10.69$ & $49.99\pm0.08$ \\
\bottomrule
\end{tabular}
\end{table}

We therefore crossed two one-time interventions at selected depths: replace the current hidden sequence with the clean Rule~30 embedding, and clear the existing CoT cache. The four paired conditions use the same examples and pre-intervention trajectory; subsequent repeats again accumulate keys and values normally. Table~\ref{tab:cotf-factorial-short} reports mean cell-accuracy changes over the next five predictions, averaged first within each run over tested intervention depths and then across runs.

\begin{table}[htbp]
\centering
\small
\caption{Five-repeat effects of CoTFormer state--cache interventions on cell accuracy, in percentage points (mean \(\pm\) sample SD across runs). State is clean-state injection with the old cache retained; reset after clean is the additional effect of clearing that cache; reset alone keeps the corrupted state. Intervention depths were selected around failure and are not balanced across policies; the paragraph on factorial effects below defines the effects.}
\label{tab:cotf-factorial-short}
\begin{tabular}{@{}lrrr@{}}
\toprule
Training cache & State & Reset after clean & Reset alone \\
\midrule
Recent 2 & \(18.2\pm16.2\) & \(31.6\pm16.6\) & \(0.1\pm0.1\) \\
Recent 3 & \(8.3\pm3.7\) & \(41.4\pm3.2\) & \(0.1\pm0.1\) \\
Recent 4 & \(3.7\pm5.7\) & \(46.2\pm5.9\) & \(0.0\pm0.0\) \\
Recent 5 & \(1.7\pm3.1\) & \(48.2\pm3.1\) & \(0.0\pm0.0\) \\
Recent 8 & \(-0.9\pm1.4\) & \(50.1\pm0.1\) & \(0.4\pm0.5\) \\
Full & \(0.2\pm0.2\) & \(49.5\pm0.3\) & \(0.0\pm0.0\) \\
\bottomrule
\end{tabular}
\end{table}

\begin{figure}[H]
\centering
\includegraphics[width=0.52\textwidth]{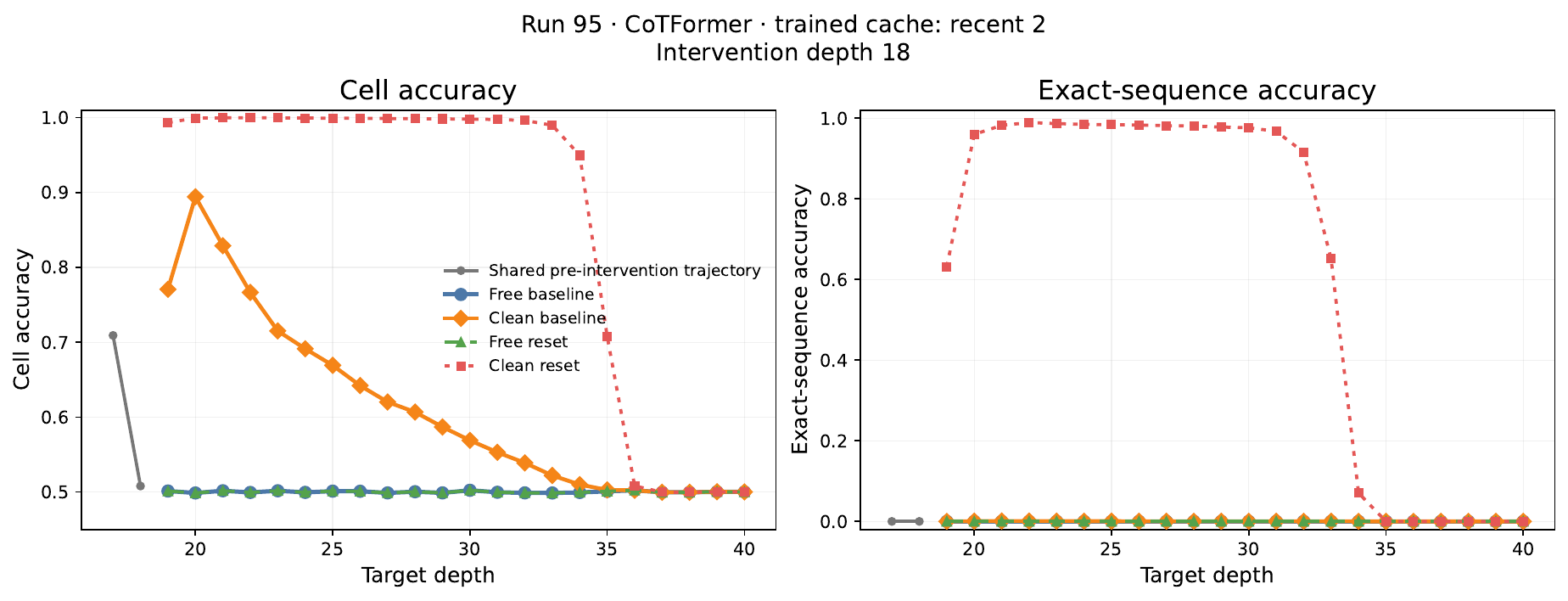}
\vspace{-0.6em}
\includegraphics[width=0.52\textwidth]{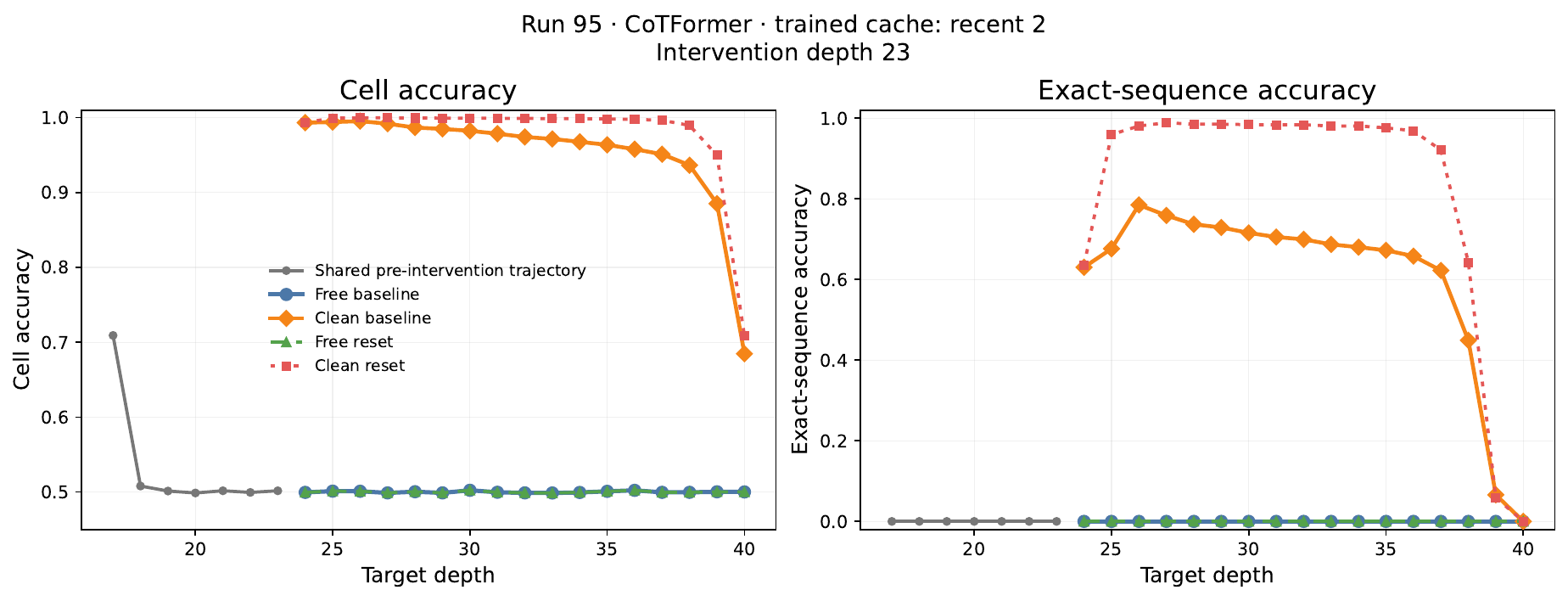}
\vspace{-0.6em}
\includegraphics[width=0.52\textwidth]{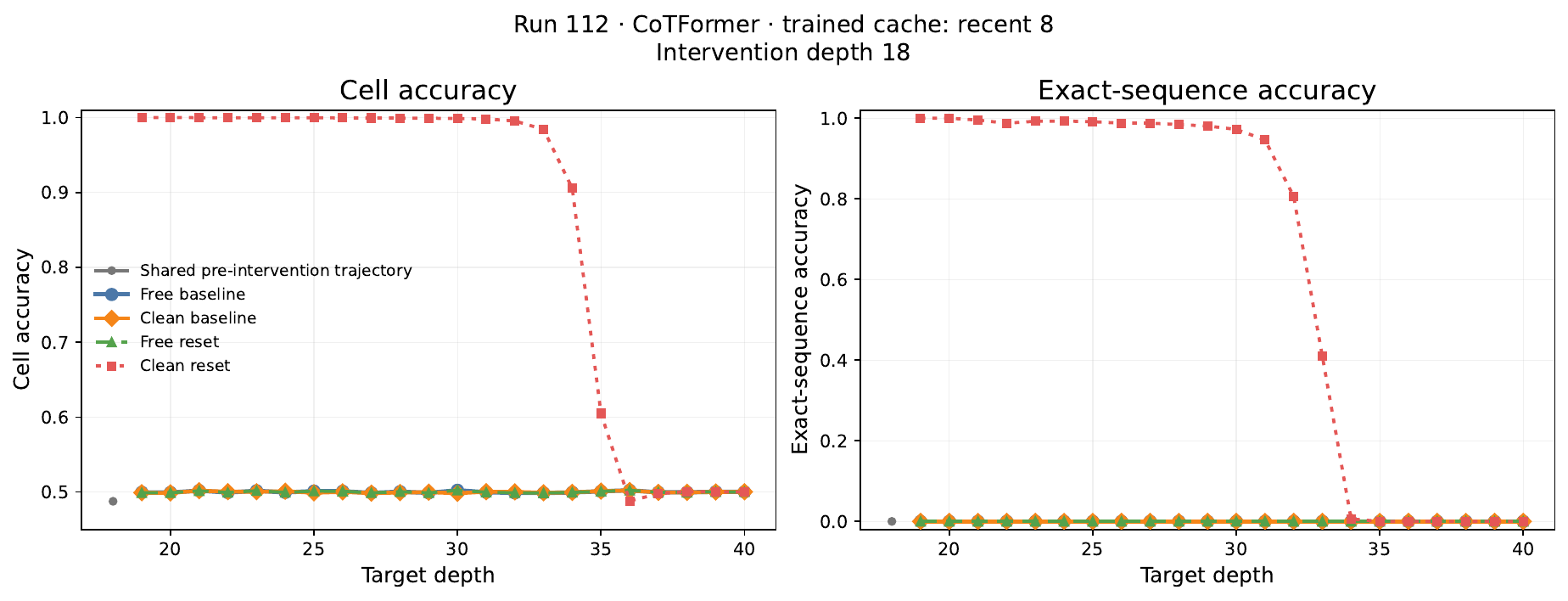}
\caption{Representative factorial state--cache interventions: recent-2 run 95 at depths 18 and 23 (top and middle), and recent-8 run 112 at depth 18 (bottom). Clean-state injection rescues the recent-2 model temporarily; the recent-8 model requires both state correction and cache reset. These are selected examples, and neither intervention makes the transition indefinitely stable.}
\label{fig:cotf-intervention-short}
\end{figure}

Correcting the state helped less as the trained cache window grew, while clearing the cache helped increasingly \emph{after} state correction. Cache reset alone had almost no effect. This pattern holds from recent-2 through recent-8: injecting a clean state and resetting
the cache together gives the most robust rescue, whereas a clean state alone rescues less robustly, temporarily for recent-2 and not at
all in the recent-8 example. Even the joint rescue is temporary, fading by depth 40. Figure~\ref{fig:cotf-intervention-short}
shows these post-intervention trajectories.

\paragraph{Factorial effects.}
\label{app:factorial}
Each intervention starts from a frozen checkpoint at source depth $t$ and runs four paired conditions on the same 4,096-row validation
split, sharing the free-running trajectory through depth $t$ and continuing to depth 40: free baseline (FB: model state and cache kept),
clean baseline (CB: hidden sequence replaced by the clean Rule~30 embedding, cache kept), free reset (FR: model state kept, cache cleared),
and clean reset (CR: both). Clearing is a one-time operation; from repeat $t+1$ onward every condition accumulates keys and values
normally under the model's trained cache policy. Writing $A_c(t,t+j)$ for cell accuracy of condition $c$ at depth $t+j$, the three
columns of Table~\ref{tab:cotf-factorial-short} are the five-repeat paired differences
\begin{align*}
\Delta_{\mathrm{state}}&=\tfrac15\textstyle\sum_{j=1}^{5}\bigl[A_{\mathrm{CB}}(t,t+j)-A_{\mathrm{FB}}(t,t+j)\bigr],\\
\Delta_{\mathrm{reset}\mid\mathrm{clean}}&=\tfrac15\textstyle\sum_{j=1}^{5}\bigl[A_{\mathrm{CR}}(t,t+j)-A_{\mathrm{CB}}(t,t+j)\bigr],\\
\Delta_{\mathrm{reset}\mid\mathrm{free}}&=\tfrac15\textstyle\sum_{j=1}^{5}\bigl[A_{\mathrm{FR}}(t,t+j)-A_{\mathrm{FB}}(t,t+j)\bigr],
\end{align*}
labelled ``State'', ``Reset after clean'', and ``Reset alone''. The total effect of the joint intervention is therefore
$\Delta_{\mathrm{state}}+\Delta_{\mathrm{reset}\mid\mathrm{clean}}$. Each run's effects are first averaged over its tested source depths, and the
table reports the mean and sample standard deviation across runs, so runs with more tested depths do not receive more weight. The rows
contain 3 runs and 16 interventions (recent 2), 2 and 10 (recent 3), 3 and 9 (recent 4), 3 and 10 (recent 5), 2 and 7 (recent 8), and 2 and
6 (full cache). One recent-8 run is excluded because its selected checkpoint was already at chance within the supervised depths, so it
had no extrapolating trajectory to rescue. No factorial grid was collected for recent-1, recent-6, or recent-7 models.

For the state--cache factorial, the first post-intervention prediction occurs at depth $t+1$. A recent-2 run that had collapsed before intervention at $t=23$ regained $99.33\%$ cell accuracy at depth 24 from state correction alone and retained $96.36\%$ at depth 35; joint state correction and cache reset reached $99.82\%$ at depth 35. A recent-8 run differed: at $t=18$, neither state correction nor reset alone helped, whereas their combination reached perfect cell and exact-row accuracy at depth 19 and $99.88\%$ cell accuracy at depth 30. Both interventions eventually failed by depth 40.

\subsection{Cache use in delayed recall}
\label{app:cotf-cache-use}
The block written during repeat \(r\) holds keys and values computed from the state entering that repeat, that is, the state after
update \(r-1\). For a query of row \(q\), block \(q\) therefore holds the state before update \(q\), and block \(q+1\) holds the
completed row. All interventions use the 25,000-step runs, each at its selected checkpoint, the one with the best validation recall
(steps 17,500 to 25,000 for these runs); those on block \(q+1\) are the basis of
Section~\ref{sec:recall-lookup}.

\paragraph{Block \(q+1\), which holds the requested row.}
We ran three interventions on block \(q+1\) of the selected checkpoints at horizons 6, 12 and 20, and on every other rollout
block for all six horizon-6 checkpoints (three seeds each with one and two recall applications, queries 1 to 5): giving the block
another input's values, hiding it, and leaving it as the only visible rollout block. Table~\ref{tab:cotf-qplus1} gives the horizon-6 results with one recall application. Two of the three seeds look the row up
at most queries (7 of their 10): giving block \(q+1\) another input's values takes recall from \(100\%\) to chance, with a Matthews correlation with the correct row of \(0.00\)--\(0.19\), hiding the block costs \(27\)--\(49\) points, and leaving it as the only visible
rollout block costs at most \(1.0\). In those cases the block receives \(25\)--\(42\%\) of the attention mass. For the other blocks,
all 72 swaps change recall by at most \(0.6\) points and 70 of 72 hides by at most \(0.6\) (the others \(2.6\) and \(4.6\)). The
third seed loses at most \(0.1\) points when block \(q+1\) is swapped or hidden and at most \(3.4\) when any single block is the
only one visible, so it answers without the cache. With two recall applications at horizon 6, one seed looks the row up in block \(q+1\)
at all five queries (swap \(-48.6\) to \(-50.1\) points) and one in block \(q+2\) at queries 1 to 4 (swap \(-49.9\) to
\(-50.1\), hide \(-47.8\) to \(-56.9\), no loss when it is the only visible block, and no effect from block \(q+1\)), so the
address is not the same in every seed. The third loses \(27\)--\(31\) points at queries 1 to 3 when block \(q+1\) is hidden
but at most \(8.4\) when any block is swapped.

At horizon 12 (queries 1, 2, 3, 6, 7, 10 and 11 at evaluation horizon 12) no well-fitted checkpoint depends on block \(q\) or \(q+1\). On the five checkpoints
with at least \(94\%\) recall (two with one recall application, three with two), swapping block \(q+1\) changes cell accuracy by at most \(1.7\) points and hiding it by at most
\(2.2\) in all 35 pairs, although queries 7, 10 and 11 are recalled at \(85\)--\(100\%\); queries 1 to 3 are recalled at
\(63\)--\(70\%\) with or without the block. The third one-application seed, which recalls \(78\%\) of cells overall, loses \(10.2\)
points at query 1 and at most \(0.4\) elsewhere. On the horizon-20 checkpoint the swap changes recall by \(-4.3\) to \(+4.9\)
points and hiding by \(-8.0\) to \(+4.9\) over eight queries. We did not measure agreement with the other input's row after a
swap, and each test touches one block, so use of several blocks together is not excluded. At horizons 12 and 20 only blocks
\(q\) and \(q+1\) were tested, so a lookup at another block, as in the seed that reads block \(q+2\), would not have been seen.

\begin{table}[htbp]
\centering
\small
\caption{CoTFormer at horizon 6 with one recall application: change in recall cell accuracy (points) when block \(q+1\), which
holds the completed requested row, is given another input's values (\emph{Swap}), hidden (\emph{Hide}), or left as the only
visible rollout block (\emph{Only}). Unmodified recall is at least \(99.6\%\) at every query.}
\label{tab:cotf-qplus1}
\begin{tabular}{@{}llrrrrr@{}}
\toprule
Seed & Intervention & \(q=1\) & \(q=2\) & \(q=3\) & \(q=4\) & \(q=5\) \\
\midrule
1 & Swap & \(-50.0\) & \(-50.0\) & \(-49.1\) & \(-42.3\) & \(-2.6\) \\
  & Hide & \(-42.1\) & \(-49.3\) & \(-40.9\) & \(-32.8\) & \(-0.1\) \\
  & Only & \(0.0\) & \(0.0\) & \(0.0\) & \(0.0\) & \(0.0\) \\
\midrule
2 & Swap & \(-47.2\) & \(-49.2\) & \(-40.4\) & \(0.0\) & \(0.0\) \\
  & Hide & \(-26.8\) & \(-45.0\) & \(-39.2\) & \(0.0\) & \(0.0\) \\
  & Only & \(-1.0\) & \(0.0\) & \(0.0\) & \(-2.9\) & \(0.0\) \\
\midrule
3 & Swap & \(0.0\) & \(0.0\) & \(0.0\) & \(0.0\) & \(0.0\) \\
  & Hide & \(0.0\) & \(+0.1\) & \(0.0\) & \(0.0\) & \(+0.1\) \\
  & Only & \(-0.5\) & \(-3.1\) & \(-0.4\) & \(-0.4\) & \(-0.9\) \\
\bottomrule
\end{tabular}
\end{table}

\paragraph{Block \(q\).}
The same checkpoints were also tested on block \(q\), called the requested block below, by permuting its values,
hiding it, or leaving it as the only visible rollout block; the baselines are identical to those of the block \(q+1\) runs.
Block \(q\) is not needed: over horizons 6, 12 and 20 together, hiding it leaves recall within \(0.6\)
points in 69 of 73 query--checkpoint pairs and permuting its values in 71 of 73.

At \(H_{\max}=6\) CoTFormer fits both the forward computation and recall, and we ran the interventions on all six horizon-6
checkpoints (three seeds each with one and two recall applications) at queries 1 to 5. Unmodified recall is \(99.6\)--\(100\%\) of
cells at every query. Permuting the requested block's values changed cell accuracy by at most \(0.09\) points in all 30 pairs.
Hiding the block changed it by at most \(0.6\) points in 28 pairs; the two exceptions are one seed with two recall applications,
which lost \(39.6\) points at query 1 and \(5.1\) at query 2, while the other two seeds of that setting lost nothing at the same
queries. Leaving only the requested block visible reduced recall by more than 40 points, to about chance, in 16 pairs, and left it
within \(0.6\) points in 7. The requested block received \(1\)--\(26\%\) of attention mass, below \(10\%\) in 23 of 30 pairs.
Changes in internal accuracy equal the changes in cell accuracy to within \(0.02\) points in every pair.

At the longer horizons, on six
checkpoints (two at horizon 12 with one recall application, three at horizon 12 with two, one at horizon 20) and 43
query--checkpoint pairs, permuting the requested block's values moved cell accuracy by more than \(0.6\) points in two pairs
(\(-1.7\) and \(-1.4\)) and hiding it in two (\(-8.9\) and \(+1.0\)); the changes in internal accuracy are the same to within \(0.5\)
points in every pair. The third one-application horizon-12 seed, whose block \(q+1\) results are given above, was tested on
block \(q\) later and is not in these counts: permuting the block costs \(8.0\) points at query 2 and hiding it \(7.4\), and
neither changes recall by more than \(0.4\) points at the other six queries. These interventions use the final-test split and all 4,096 rows.

\section{Further delayed-recall results}
\label{app:recall-further}
Table~\ref{tab:recall-25k-k2} gives the results with two recall applications, and Table~\ref{tab:recall-internal} the internal
recall measures and the unique-best rate.

\begin{table}[htbp]
\centering
\small
\caption{Delayed recall after 25,000 steps with two recall applications, for the four models trained in that setting: cell / exact-row
accuracy (\%), mean \(\pm\) sample SD over three seeds.}
\label{tab:recall-25k-k2}
\begin{tabular}{@{}lrrrr@{}}
\toprule
 & \multicolumn{2}{c}{\(H_{\max}=12\)} & \multicolumn{2}{c}{\(H_{\max}=20\)} \\
\cmidrule(lr){2-3}\cmidrule(l){4-5}
Model & Cell & Exact row & Cell & Exact row \\
\midrule
Gated BUT & \(99.55\pm0.27\) & \(96.36\pm2.28\) & \(94.90\pm1.24\) & \(76.07\pm4.24\) \\
LSTM-UT & \(98.68\pm0.23\) & \(89.82\pm3.52\) & \(92.66\pm1.64\) & \(63.96\pm16.93\) \\
BUT & \(96.08\pm1.53\) & \(75.48\pm5.99\) & \(88.16\pm1.62\) & \(49.60\pm3.71\) \\
CoTFormer & \(95.13\pm0.25\) & \(65.75\pm2.86\) & \(84.85\pm1.28\) & \(33.35\pm7.83\) \\
\bottomrule
\end{tabular}
\end{table}

\begin{table}[htbp]
\centering
\small
\caption{Internal recall accuracy and unique-best rates, after 25,000 steps with one recall application (\%, mean \(\pm\)
sample SD over three seeds). \emph{Internal}: cell accuracy against the model's own decoded row at the requested repeat.
\emph{Changed}: the same on cells where the requested and current decoded rows differ. \emph{Unique best}: rate at which the answer
is uniquely closest to the true requested row among all true rows of the trajectory. \emph{Cosine best}: rate at which the answer's
logits are uniquely closest in cosine similarity to the model's own logits at the requested repeat, among its logits at every
rollout repeat.}
\label{tab:recall-internal}
\setlength{\tabcolsep}{3pt}
\resizebox{\textwidth}{!}{\begin{tabular}{@{}lrrrrrrrr@{}}
\toprule
 & \multicolumn{4}{c}{\(H_{\max}=12\)} & \multicolumn{4}{c}{\(H_{\max}=20\)} \\
\cmidrule(lr){2-5}\cmidrule(l){6-9}
Model & Internal & Changed & Unique best & Cosine best & Internal & Changed & Unique best & Cosine best \\
\midrule
Gated BUT & \(97.24\pm2.82\) & \(97.16\pm2.69\) & \(95.52\pm3.86\) & \(95.43\pm4.64\) & \(94.53\pm1.21\) & \(94.27\pm1.52\) & \(89.52\pm3.12\) & \(90.12\pm2.89\) \\
LSTM-UT & \(97.95\pm0.19\) & \(97.91\pm0.22\) & \(97.44\pm1.07\) & \(98.36\pm0.69\) & \(92.41\pm1.00\) & \(92.46\pm1.03\) & \(84.55\pm0.75\) & \(85.99\pm0.78\) \\
BUT & \(96.17\pm0.10\) & \(96.20\pm0.40\) & \(92.71\pm1.50\) & \(93.33\pm2.14\) & \(89.00\pm2.51\) & \(89.61\pm1.94\) & \(78.49\pm6.21\) & \(80.10\pm5.84\) \\
Decay BUT, \(\lambda=0.9\) & \(96.27\pm0.46\) & \(96.18\pm0.56\) & \(93.85\pm0.55\) & \(94.33\pm0.82\) & \(89.98\pm0.54\) & \(89.97\pm0.61\) & \(81.77\pm2.07\) & \(83.57\pm2.77\) \\
Decay BUT, \(\lambda=0.5\) & \(96.47\pm0.71\) & \(96.70\pm0.69\) & \(93.51\pm2.46\) & \(92.86\pm4.26\) & \(88.55\pm0.28\) & \(88.62\pm1.07\) & \(78.85\pm2.25\) & \(80.18\pm3.47\) \\
Linear memory & \(96.20\pm0.33\) & \(95.90\pm0.57\) & \(93.14\pm2.07\) & \(94.17\pm1.96\) & \(88.43\pm3.83\) & \(88.55\pm3.80\) & \(79.11\pm8.62\) & \(81.69\pm6.16\) \\
FPRM BUT & \(95.07\pm1.75\) & \(94.95\pm1.86\) & \(93.61\pm2.76\) & \(94.32\pm2.08\) & \(86.03\pm3.45\) & \(85.91\pm3.19\) & \(75.00\pm8.30\) & \(77.39\pm6.80\) \\
Gated decay BUT, \(\lambda=0.9\) & \(97.26\pm1.34\) & \(97.16\pm1.38\) & \(97.24\pm2.00\) & \(97.14\pm2.24\) & \(90.01\pm2.69\) & \(89.95\pm2.37\) & \(81.58\pm6.30\) & \(82.83\pm6.04\) \\
CoTFormer & \(90.05\pm7.88\) & \(90.19\pm7.89\) & \(79.35\pm16.68\) & \(82.52\pm14.87\) & \(86.76\pm6.89\) & \(86.02\pm7.48\) & \(74.32\pm12.96\) & \(75.89\pm11.75\) \\
\bottomrule
\end{tabular}}
\end{table}

\subsection{Protocol details}
\label{app:recall-protocol}
BUT must retain or reconstruct the past from its current hidden sequence; CoTFormer can additionally attend to cached keys and
values from earlier repeats; LSTM-UT can additionally read its bounded cell state, which persists through the recall applications.
Linear memory carries its cell from evolution into recall, as LSTM-UT does. All models receive the same learned direction and age
controller. The loss is \(\mathcal L=(\mathcal L_{\rm evo}+2\mathcal L_{\rm query})/3\); the evolution loss averages evenly over
the \(H\) repeats. Horizons \(1{:}1,\ldots,H_{\max}{:}H_{\max}\) and their queries cycle deterministically. The main sweep uses
three model-initialisation seeds, a common data seed, 64-cell rows, width 64, four heads, and the same \(0/1/0\) shared-block
architecture for all models. Training has 100,000 initial rows; validation and test each have 4,096 independent rows. The selected
checkpoint is the one with the highest ground-truth recall cell accuracy on validation, averaged first over nontrivial queries
within each horizon and then equally over horizons. All reported main-sweep results use frozen checkpoints on the final-test split.
The architectures use identical data, splits, optimiser settings, training length, and checkpoint candidates, differing only in
architecture and model seed.

\subsection{Notes on the 25,000-step comparison}
\label{app:recall-notes}
\paragraph{Forward-pass errors.}
A model that computes the rows imperfectly could lose recall accuracy without any failure of memory. Internal accuracy removes that
effect, and it differs from cell accuracy by at most \(1.0\) point for every model at both horizons (Table~\ref{tab:recall-25k}).
Restricting it to the cells on which the requested and current rows differ, so that returning the current row earns nothing, changes
it by at most a further \(0.8\) points (Appendix Table~\ref{tab:recall-internal}). None of the comparisons we draw changes on either measure.

\paragraph{Seed dependence.}
Table~\ref{tab:recall-per-seed} gives every horizon-20 seed. No family fits the rollout equally in all three seeds: the share of
final rows computed correctly at depth 20 spans at least 11 points in every family that fits, including LSTM-UT (\(80\)--\(100\%\))
and gated BUT (\(77\)--\(99\%\)). The seeds that fit worst are the three CoTFormer seeds, one decay BUT seed with
\(\lambda=0.5\) that computes almost no rows past depth 8, and one FPRM BUT seed. Gating does not separate the fits: BUT and decay
BUT with \(\lambda=0.9\) fit as well as gated BUT, and gated decay BUT fits worse than both. Recall is averaged over horizons 1 to
20, so a seed that fails at depth 20 can still recall rows at shorter horizons.

\begin{table}[htbp]
\centering
\footnotesize
\caption{Per-seed results at \(H_{\max}=20\) after 25,000 steps with one recall application, at the selected checkpoint (final test, \%).
\emph{Rows computed}: exact-row accuracy of the forward computation, as the mean over rollout depths 1 to 20 and at depth 20.
\emph{Recall}: cell and exact-row accuracy against the true requested row, as in Table~\ref{tab:recall-25k}, whose means these seeds give.}
\label{tab:recall-per-seed}
\begin{tabular}{@{}lcrrrr@{}}
\toprule
 & & \multicolumn{2}{c}{Rows computed} & \multicolumn{2}{c}{Recall} \\
\cmidrule(lr){3-4}\cmidrule(l){5-6}
Model & Seed & Depths 1--20 & Depth 20 & Cell & Exact row \\
\midrule
Gated BUT & 1 & \(92.11\) & \(77.29\) & \(95.82\) & \(71.62\) \\
 & 2 & \(99.49\) & \(99.10\) & \(94.35\) & \(73.92\) \\
 & 3 & \(95.55\) & \(90.55\) & \(93.42\) & \(65.84\) \\
\addlinespace[2pt]
LSTM-UT & 1 & \(100.00\) & \(99.95\) & \(93.31\) & \(69.97\) \\
 & 2 & \(98.95\) & \(97.88\) & \(92.59\) & \(68.42\) \\
 & 3 & \(89.24\) & \(79.98\) & \(91.34\) & \(64.53\) \\
\addlinespace[2pt]
BUT & 1 & \(97.71\) & \(90.50\) & \(91.01\) & \(57.67\) \\
 & 2 & \(97.03\) & \(92.43\) & \(89.79\) & \(54.72\) \\
 & 3 & \(96.56\) & \(73.17\) & \(86.10\) & \(48.85\) \\
\addlinespace[2pt]
Decay BUT, \(\lambda=0.9\) & 1 & \(97.97\) & \(94.87\) & \(90.47\) & \(63.49\) \\
 & 2 & \(95.22\) & \(87.40\) & \(90.05\) & \(49.63\) \\
 & 3 & \(97.07\) & \(83.57\) & \(89.40\) & \(48.02\) \\
\addlinespace[2pt]
Decay BUT, \(\lambda=0.5\) & 1 & \(21.41\) & \(0.05\) & \(87.04\) & \(42.52\) \\
 & 2 & \(97.42\) & \(95.36\) & \(88.80\) & \(49.31\) \\
 & 3 & \(90.13\) & \(81.18\) & \(88.55\) & \(42.17\) \\
\addlinespace[2pt]
Linear memory & 1 & \(85.29\) & \(58.15\) & \(84.42\) & \(36.11\) \\
 & 2 & \(88.21\) & \(79.93\) & \(88.75\) & \(56.08\) \\
 & 3 & \(98.09\) & \(93.68\) & \(92.12\) & \(56.96\) \\
\addlinespace[2pt]
FPRM BUT & 1 & \(99.39\) & \(98.19\) & \(82.91\) & \(33.54\) \\
 & 2 & \(96.25\) & \(92.75\) & \(89.74\) & \(46.40\) \\
 & 3 & \(60.91\) & \(41.97\) & \(85.38\) & \(40.04\) \\
\addlinespace[2pt]
Gated decay BUT, \(\lambda=0.9\) & 1 & \(94.57\) & \(90.87\) & \(93.12\) & \(64.87\) \\
 & 2 & \(75.35\) & \(66.46\) & \(88.55\) & \(40.61\) \\
 & 3 & \(89.00\) & \(84.35\) & \(88.31\) & \(53.57\) \\
\addlinespace[2pt]
CoTFormer & 1 & \(65.26\) & \(0.00\) & \(94.30\) & \(64.46\) \\
 & 2 & \(18.77\) & \(0.00\) & \(83.47\) & \(18.90\) \\
 & 3 & \(12.44\) & \(0.00\) & \(80.81\) & \(13.11\) \\
\bottomrule
\end{tabular}
\end{table}

\paragraph{Recall by age.}
After 20 steps, internal cell accuracy stays above \(95\%\) up to age 7 for LSTM-UT and age 9 for gated BUT, against ages 4 to 5
for BUT and decay BUT. At age 8 gated BUT still reproduces \(94\%\) of whole rows, where BUT and the two decay BUTs reproduce none.
From age 13 the four models of Figure~\ref{fig:recall-age} are at \(64\%\) of cells or below, against \(50\%\) for a guess, and
at age 19 the seven models other than CoTFormer and gated decay BUT are all at \(53\)--\(57\%\). The CoTFormer checkpoints examined in
Section~\ref{sec:recall-lookup} recall the most recent rows at \(97\)--\(100\%\) and the oldest at \(55\)--\(69\%\).

\paragraph{CoTFormer's fit.}
CoTFormer stores every earlier state and recalls no better than BUT. It is last in exact rows at both horizons, it is below BUT on
internal accuracy as well, and its seeds vary most (SD of 28 to 31 exact-row points). Its forward computation is the least fitted: at the
last trained depth it gets \(27.6\%\) of rows exactly right at horizon 12 and \(0.0\%\) at horizon 20, against \(68\)--\(93\%\) for
BUT, LSTM-UT and gated BUT.

\paragraph{LSTM-UT's lead over BUT.}
At horizon 12 with one recall application LSTM-UT leads BUT by \(1.8\) cell points; in exact rows the lead is \(9.5\) points at
horizon 12 and \(13.9\) at horizon 20. Most selected
checkpoints are at or near the final step, so these runs show the ranking under equal training and not at convergence. FPRM BUT without input re-injection failed
to train on this task in two of three seeds at horizons 12 and 20, so it is not compared. With re-injection FPRM could recompute the
first row from the input instead of recalling it, but it does not: at horizon 12 it scores \(58.6\%\) on that query.

\paragraph{Decay strength and the age of the recalled row.}
For decay BUT at horizon 20, cell accuracy by the age of the requested row is at least \(94\%\) up to age 6 with \(\lambda=0.9\) and
falls to \(54.8\%\) at age 19. With \(\lambda=0.5\) the oldest rows are recalled less well: at ages 12 to 19 every seed is below every
\(\lambda=0.9\) seed, and the mean is \(2.0\) to \(6.3\) cell points lower. At younger ages the \(\lambda=0.5\) seeds spread widely,
because one of them gets only \(65\%\) of the cells of the current row, so its recall there is limited by its forward pass.

\begin{figure}[htbp]
\centering
\includegraphics[width=\textwidth]{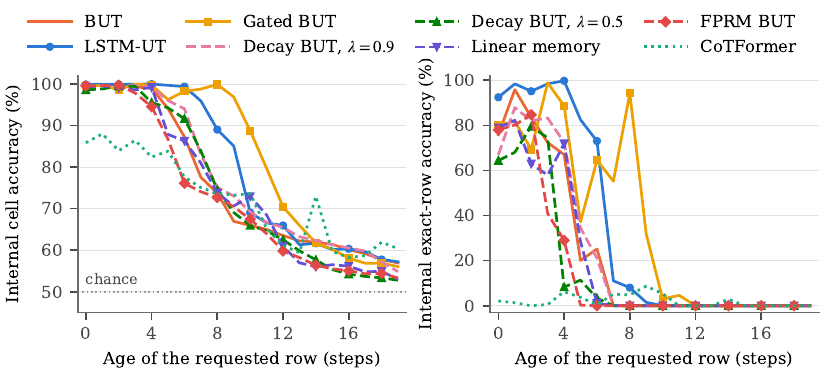}
\caption{Recall by age after a 20-step rollout for all eight models, as in Figure~\ref{fig:recall-age}. Means of three seeds;
seed ranges are omitted because they overlap. CoTFormer's exact-row agreement is low at every age, including age 0, the current row.}
\label{fig:recall-age-all}
\end{figure}

\subsection{LSTM-UT cell interventions}
\label{app:cell-interventions}
\begin{figure}[htbp]
\centering
\includegraphics[width=\textwidth]{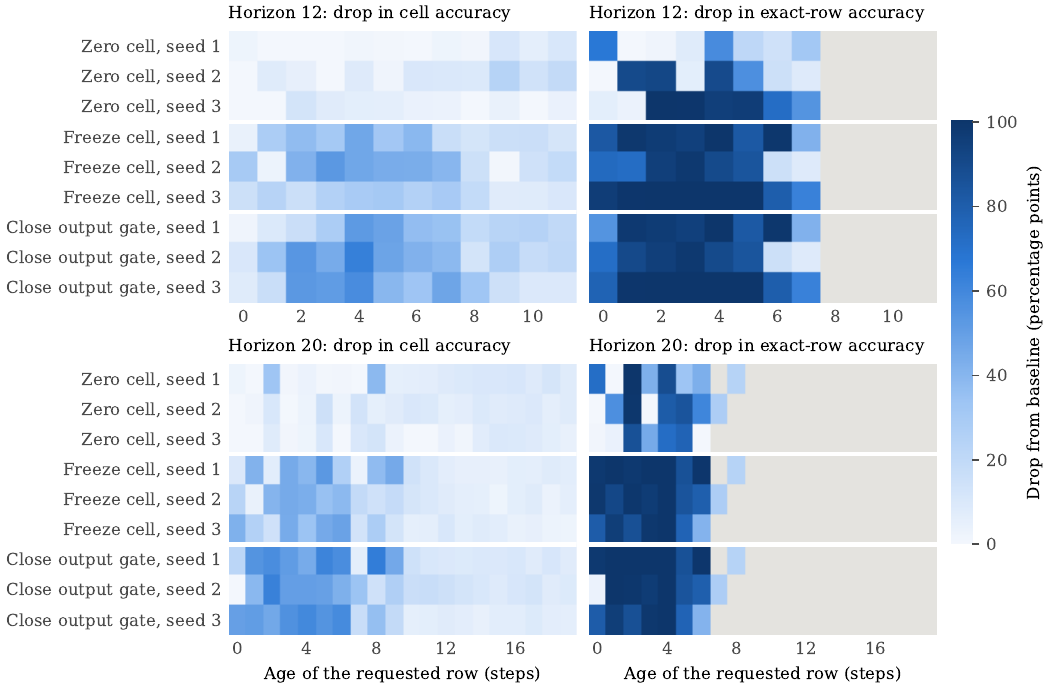}
\caption{Drop in internal recall accuracy from baseline under three cell interventions, by age of the requested row, for each
selected LSTM-UT checkpoint (25,000-step runs) with one recall application. Left: cell accuracy. Right: exact-row accuracy; ages where the baseline
has under \(5\%\) exact rows are grey.}
\label{fig:cell-drop}
\end{figure}

All interventions act only during the recall application; the rollout is unmodified. \emph{Zero cell} sets the cell to zero before
the recall application, which then rewrites and reads it as usual. \emph{Freeze cell} skips the cell update, so the stream receives
the end-of-rollout cell without the forget gate or the step's own write. \emph{Close output gate} updates the cell but does not read
it. \emph{Swap cell} gives each input the end-of-rollout cell of the previous input in its batch of 128; test rows are independent
random draws. Figure~\ref{fig:cell-drop} shows, for each of the six selected checkpoints, the drop in internal accuracy at each
age under the first three. Averaged as in Table~\ref{tab:recall-25k}, freezing leaves \(72.7\%\) and \(67.2\%\) of cells and
\(6.3\%\) and \(3.2\%\) of exact rows at horizons 12 and 20, and closing the output gate \(62.7\%\) and \(51.2\%\) of cells and
\(3.3\%\) and \(0.2\%\) of exact rows. Internal accuracy differs from accuracy against the true row by at most \(0.1\) points in
every condition. Setting the cell to zero costs \(2.8\) and \(5.6\) cell points on average at horizons 12 and 20, and the loss falls on ages that differ
between seeds. Freezing the cell or closing the output gate also breaks the current-row query, which needs no memory (\(75\)--\(93\%\) of
cells), and so does not isolate memory.

For the swap, the measures against the donor's true row agree with Figure~\ref{fig:cell-swap}. Two unrelated rows differ in 32 of
64 cells on average, and without the swap the output is 32 cells from the donor's row at every age. At the age each checkpoint
follows the donor most, the mean distance to the donor's row falls to \(0.0\), \(4.0\), \(7.3\), \(11.7\), \(14.8\) and
\(20.3\) cells, and the Matthews correlation with the donor's row rises from \(0\) to \(1.00\), \(0.88\), \(0.78\), \(0.65\),
\(0.54\) and \(0.37\).

The donor share of Figure~\ref{fig:cell-swap} is counted on the cells where the input's own answer and the donor's differ, about
half of the 64; \(50\%\) means the output says nothing about either row. Among the ages a checkpoint recalls well, the highest donor
share is \(37\)--\(100\%\) across the six checkpoints, against at most \(5\%\) without the swap; it exceeds \(10\%\) at two to five of
these ages and is \(0\)--\(7\%\) at the rest. Where the two answers agree, accuracy under the swap is at most \(4.5\) points
below baseline in five checkpoints and \(9.4\) in the sixth: the foreign cell is read, and does not merely disrupt the step.

\section{Full ablation results}
\label{app:ablations}
This appendix gives the full version of Section~\ref{sec:ablations}. It adds a \emph{linear read} variant of LSTM-UT, capacity
controls, per-seed results and further checks. Rates called \emph{shift-quotiented} are the leading Lyapunov rates of
Section~\ref{sec:abl-measure}, measured after removing per-token offsets.

\begin{table}[htbp]
\centering
\small
\caption{The models, by what is carried between repeats and whether it decays; a gated decay is set by the input. \(^\dagger\) marks input-dependent gates; in gated decay BUT the gate acts on the write and the decay is fixed. We trained
no model whose only carried state is a stream with an input-dependent decay.}
\label{tab:models}
\setlength{\tabcolsep}{4pt}
\footnotesize
\resizebox{\textwidth}{!}{\begin{tabular}{@{}llll@{}}
\toprule
Carried state & No decay & Fixed decay & Gated decay \\
\midrule
Stream & BUT, BUT + injection, gated BUT\(^\dagger\) & decay BUT, FPRM BUT, gated decay BUT\(^\dagger\) & not tested \\
Stream + side state & No-forget LSTM-UT\(^\dagger\) & linear memory & LSTM-UT\(^\dagger\) \\
Stream + earlier states & CoTFormer & & \\
\bottomrule
\end{tabular}}
\end{table}

LSTM-UT changes BUT in three ways at once: it carries a second state, the cell \(c\), from one repeat to the next; \(c\) decays
through a forget gate and is written through an input gate; and it is added back to the residual stream through an output gate and a
\(\tanh\). It also adds parameters. Any of these could explain why it extrapolates. We separate them with variants that each remove one
component, replace it with a simpler one, or add a known alternative. Every variant uses the \(R_{12}\) protocol of
Section~\ref{sec:rule30} (trained on depths 1--12, selected at depth 15, evaluated to depth 60), five seeds, and the same checkpoint
selection. Unless stated otherwise, numbers are final-test results at the validation-selected unconstrained extrapolation checkpoint,
and \(\pm\) is the sample standard deviation over seeds.

\subsection{Measurement definitions}
\label{app:abl-measure}
Each diagnostic is a rollout to repeat 60 on the same 64 rows, with every repeat decoded, compared with the unmodified rollout. One
direction is excluded throughout: adding a constant to every feature of a token is removed by every LayerNorm, so it never reaches a
decoded bit, yet a weight-1 stream carries it forever. All perturbations of \(x\) therefore have zero mean over each token's features.

\emph{Leading Lyapunov rate.} The rollout is unchanged. We carry 32 orthonormal tangent vectors beside it. At every repeat they are
multiplied by the Jacobian of that repeat (forward-mode Jacobian--vector products), their per-token mean is subtracted, and a QR
decomposition re-orthonormalises them (Benettin's method). The log of the first diagonal entry of \(R\), averaged over a window of
repeats and over rows, is the leading rate; a negative rate means infinitesimal errors shrink along the model's own trajectory.

\emph{Rotations.} At one repeat we rotate each token's \(x\) by an angle \(\theta\) towards a random direction orthogonal to the
all-ones vector and to the token's centred \(x\). Mean and norm are unchanged, so \(\theta\) is the angle seen by every LayerNorm and
is comparable across models. The unmodified model then runs to repeat 60. A small rotation (\(0.01\)\,rad at repeat 13) leaves the
rows exact; the distance between the perturbed and clean \(x\) over repeats 13--20 gives a rate and a halving time, the same at
\(0.001\) and \(0.1\)\,rad. A large rotation (\(0.6\) or \(1.0\)\,rad at repeat 6 or 13; 512 branches per seed) eventually flips
bits. This is akin to perturbing a trajectory, and tests whether a model recovers from errors, returning to valid Rule~30 computation, or falls into a stale non-benign attractor.

\emph{Faithful computation.} Rule~30 is left-permutive: a difference in the rightmost wrong cell always changes its right neighbour at
the next step, so a wrong row never becomes the true row again, and the best a rollout can do is to keep evolving the wrong row
correctly. A repeat is \emph{faithful} when every bit equals Rule~30 applied to the model's own previous row and the row is
neither constant nor frozen. A broken branch \emph{recovers} if it returns to five consecutive faithful repeats within 15, counted
only on rows where the unperturbed model stays faithful.

\emph{Gate closure.} In the cell models a repeat is the block update, the cell write, then the read that adds the cell to \(x\). We
skip the read for one repeat (at repeat 6, 13, 20 or 30), then run the unmodified model to repeat 60.

\subsection{Which ingredient matters}
\label{app:abl-ingredient}

\paragraph{Removing parts of LSTM-UT.}
The \emph{linear read} replaces the read \(\bar h+o\odot\tanh(c^+)\) by \(\bar h+o\odot c^+\). Only the saturation of the read is removed; gates,
cell update and parameter count are unchanged. The cell is then a gated, input-dependent linear recurrence across depth read out
linearly, the structure LSTM-UT shares with selective state-space models such as Mamba \citep{gu2023mamba}. \emph{No forget gate} sets
\(c^+=c+i\odot\widetilde c\): the cell becomes an integrator that never forgets, and only the read stays bounded. \emph{Linear
memory} is BUT plus a separate cell with no gates and no \(\tanh\), \(c^+=\lambda\odot c+W_x\hat x+W_hh\) and \(x^+=\bar h+c^+\), where \(\bar h\) is the BUT block's output, \(\hat x\) and \(h\) are the
normalised input and output of Section~\ref{sec:architecture}, and \(\lambda\) is one learned constant per channel. It keeps only a decaying side state added to the stream.

\paragraph{Changing BUT's stream instead of adding a cell.}
\emph{Decay BUT} computes \(x^+=\lambda x+\Delta(x)\), where \(\Delta\) is the block's update (attention and MLP) and
\(\lambda\) is a fixed constant, 0.9 or 0.5; BUT is \(\lambda=1\). Nothing else changes: same block, same parameters (49,472), same
training. \emph{FPRM BUT} \citep{movahedi2026fixedpoint} uses learned per-channel residual scaling that keeps about \(14\%\) of the
stream per repeat (\(\alpha_2\alpha_1^2\approx0.14\)) and re-adds the embedded input \(e\) with weight \(\beta_2\) every repeat, starting from
\(x=e\). \emph{FPRM without re-injection} is the same without the \(+\beta_2 e\) term; the input enters only as the initial stream.

\paragraph{Alternatives that do not decay.}
\emph{Gated BUT} \citep{hegazy2026gated} computes \(x^+=G\odot\bar h+(1-G)\odot x\) with LSTM-UT's gate inputs:
gating across depth without a side state. \emph{BUT with input injection} computes \(x^+=\mathcal T_\theta(x+e)\) after a first repeat \(\mathcal T_\theta(e)\), which is the standard remedy for depth extrapolation in looped models
\citep{bansal2022overthinking,geiping2026hugin,yang2024loopedlearning,lai2026testtime}.

\paragraph{A gate and a fixed decay without a cell.}
\emph{Gated decay BUT} adds decay BUT's fixed \(\lambda=0.9\) to gated BUT: \(x^+=\lambda x+G\odot\Delta(x)\), where
\(\Delta(x)=\bar h-x\). The gate scales what is written and \(\lambda\) what is kept; the parameters are gated BUT's (57,792).
In LSTM terms it has a learned input gate and a constant forget gate.

\paragraph{Capacity controls.}
BUT at width 80 (about LSTM-UT's parameter count), BUT at width 96 with four heads (more than LSTM-UT), and LSTM-UT at width 48 (about
BUT's). CoTFormer is included for completeness; Appendix~\ref{app:cotformer} gives its results in full.

\begin{table}[htbp]
\centering
\footnotesize
\caption{Depth extrapolation of every variant under \(R_{12}\) (trained on depths 1--12): final-test exact-row accuracy (\%), mean
\(\pm\) sample SD over five seeds. \emph{Reach} is, per seed (seeds 0--4 in order), the furthest depth up to which exact-row accuracy
stays at or above \(95\%\) at every depth from 12 on. Evaluation stops at depth 60, so a reach of 60 means at least 60; ``--'' marks a
seed below \(95\%\) at depth 12, which has no reach. Parameters are non-embedding; the two marked \(\approx\) are rounded from the
training logs.}
\label{tab:app-abl-extrapolation}
\setlength{\tabcolsep}{3pt}
\begin{tabular}{@{}lrrrrrl@{}}
\toprule
Model & Width & Params & Depth 20 & Depth 30 & Depth 60 & Reach per seed \\
\midrule
\multicolumn{7}{@{}l}{\emph{Decaying side state}} \\
LSTM-UT & 64 & 82,368 & \(100.00\pm0.01\) & \(99.95\pm0.07\) & \(99.67\pm0.56\) & 60, 60, 60, 60, 60 \\
Linear read & 64 & 82,368 & \(99.99\pm0.02\) & \(99.95\pm0.06\) & \(99.75\pm0.35\) & 60, 60, 60, 60, 60 \\
Linear memory & 64 & 57,856 & \(99.99\pm0.02\) & \(99.98\pm0.02\) & \(99.98\pm0.03\) & 60, 60, 60, 60, 60 \\
\multicolumn{7}{@{}l}{\emph{Decaying stream}} \\
Decay BUT, \(\lambda=0.9\) & 64 & 49,472 & \(99.98\pm0.03\) & \(99.96\pm0.04\) & \(99.56\pm0.53\) & 60, 60, 60, 60, 60 \\
Decay BUT, \(\lambda=0.5\) & 64 & 49,472 & \(99.99\pm0.03\) & \(99.99\pm0.03\) & \(99.98\pm0.04\) & 60, 60, 60, 60, 60 \\
FPRM BUT & 64 & 49,600 & \(99.97\pm0.04\) & \(99.97\pm0.05\) & \(99.95\pm0.08\) & 60, 60, 60, 60, 60 \\
FPRM, no re-injection & 64 & 49,600 & \(100.00\pm0.00\) & \(100.00\pm0.00\) & \(100.00\pm0.00\) & 60, 60, 60, 60, 60 \\
\multicolumn{7}{@{}l}{\emph{Decaying gated stream} (\(\lambda=0.9\))} \\
Gated decay BUT, 0.9 & 64 & 57,792 & \(89.72\pm22.13\) & \(69.99\pm40.95\) & \(41.25\pm51.79\) & --, 60, 24, 24, 60 \\
\multicolumn{7}{@{}l}{\emph{Side state that never decays}} \\
No forget gate & 64 & 74,176 & \(98.58\pm2.99\) & \(77.80\pm43.73\) & \(43.17\pm50.27\) & 35, 60, 27, 60, 19 \\
\multicolumn{7}{@{}l}{\emph{No decay, no side state (undecayed stream)}} \\
Gated BUT & 64 & 57,792 & \(97.90\pm2.19\) & \(4.12\pm9.20\) & \(0.00\pm0.00\) & 23, 20, 20, 19, 23 \\
BUT + injection & 64 & 49,472 & \(97.04\pm3.19\) & \(25.41\pm33.51\) & \(0.00\pm0.00\) & 20, 25, 23, 19, 21 \\
BUT & 64 & 49,472 & \(93.15\pm11.63\) & \(28.77\pm41.66\) & \(0.00\pm0.00\) & 23, 18, 24, 19, 31 \\
BUT & 80 & 77,200 & \(92.96\pm8.83\) & \(16.39\pm36.45\) & \(0.00\pm0.00\) & 21, 24, 20, 19, 14 \\
BUT, 4 heads & 96 & \(\approx\)0.11\,M & \(72.98\pm41.81\) & \(22.15\pm30.98\) & \(0.00\pm0.00\) & 24, 18, 27, 18, 16 \\
\multicolumn{7}{@{}l}{\emph{Capacity control and history cache}} \\
LSTM-UT & 48 & 46,416 & \(53.59\pm49.86\) & \(49.26\pm46.29\) & \(38.51\pm44.51\) & 60, --, 18, --, -- \\
CoTFormer & 64 & \(\approx\)0.05\,M & \(0.00\pm0.00\) & \(0.00\pm0.00\) & \(0.00\pm0.00\) & 13, --, --, --, -- \\
\bottomrule
\end{tabular}
\end{table}

Table~\ref{tab:app-abl-extrapolation} splits on one property, with one exception: whether some state carried from repeat to repeat is multiplied by
less than one. All 35 seeds of seven families with a decaying state reach depth 60. The decay can sit in a side state that is added
to the stream (LSTM-UT, linear read, linear memory) or in the stream itself (decay BUT, FPRM with and without re-injection). All 25
seeds of the five undecayed-stream families stop between depths 14 and 31, whatever their width (64, 80, 96), gating or input
injection. The one family with a side state that never decays, no forget gate, is seed dependent, with reach 19, 27, 35, 60 and 60.
Cell accuracy at depth 60 is \(99.92\)--\(100.00\%\) for each of those seven families, \(83.59\%\) without the forget gate, and
\(49.98\)--\(50.05\%\), chance, for every undecayed-stream family and CoTFormer.

The exception is gated decay BUT, an eighth decaying family. Seed 1 is at \(100.00\%\) exact rows at every depth to 60 and seed 4 declines slowly to
\(95.73\%\). Seeds 2 and 3 fit, stay above \(99\%\) at depth 20, and fall to \(8.18\%\) and \(2.34\%\) at depth 60, with \(71.28\%\) and \(64.75\%\)
of cells correct. Seed 0 has \(94.38\%\) at depth 12 and \(0\%\) at depth 30; it is included in every mean we report for this family. Over the four seeds that fit,
depth 60 is \(51.56\pm53.55\%\), the late rate \(+0.029\pm0.128\), the halving time 7 repeats and recovery \(42.8\%\), so leaving it
out changes no conclusion. Selecting in distribution gives the same reach in every
seed and changes depth 60 in one (seed 2, \(45.73\%\)). The outcome depends on when training stops: in seeds 2 and 3 another checkpoint
of the same run (steps 10,500 and 20,000) has \(100.0\%\) and \(97.9\%\) exact rows at depth 60. The protocol's rule, which looks only at depth 15, does not
pick them, and we report what it selects. A fixed decay therefore makes gated BUT's stream contract (Section~\ref{app:abl-contraction}) without making its rollout reliable.

This rules out four explanations. \emph{Read saturation is not needed}: the linear read and linear memory have no \(\tanh\) on the read
and match LSTM-UT. \emph{Gates are not needed}: linear memory has no gates and decay BUT has nothing learned in its decay, yet both reach
60 in every seed; gating without decay does not help, since gated BUT's reach (19--23) is BUT's (18--31). \emph{Re-anchoring to the
input is neither needed nor sufficient}: BUT with input injection reaches 19--25, and FPRM extrapolates with or without re-injection.
In a rollout the target at step \(t\) depends on the current row, not on the input, so re-reading the input gives no correcting
signal. \emph{Parameters do not explain it}: decay BUT has exactly BUT's 49,472 parameters, and BUT at 77,200 or about 110,000
parameters does not extend its reach (14--27). LSTM-UT at width 48 is hard to train: two seeds do not fit the trained depths, one fits
them partly, and of the three that fit at least partly, one is exact to depth 60, one stays near \(70\%\) from depth 20 to 60, and one
has a reach of 18. CoTFormer does not fit the trained depths in four of five seeds under \(R_{12}\), so its row reflects a fitting
failure rather than an extrapolation result; Appendix~\ref{app:cotformer} gives the reach of every fitted CoTFormer run.

Without the forget gate, a failing seed keeps balanced rows and \(83.6\%\) correct cells at depth 60, whereas the undecayed-stream
models end at chance. Section~\ref{app:abl-decay} shows directly
that decay is sufficient. The checkpoint selector does not change the picture: under in-distribution selection each of the seven decaying families that reach depth 60
still does so in every seed except one linear-read seed (reach 57; depth-60 exact-row accuracy \(98.80\pm2.51\%\)), and every
undecayed-stream family still ends at \(0\%\).

Table~\ref{tab:ones-rate} gives the share of predicted ones for every $R_{12}$ family. Rows of the true automaton have $49.7$--$50.2\%$ ones at depths 13--60. Every seed of the six decaying families that reach depth 60 stays within $0.4$ points of $50\%$; every BUT and gated BUT seed leaves the $25$--$75\%$ band, towards all ones or all zeros. Gated decay BUT does not fail that way: its two fitted seeds that stop at depth 24 end at $49.3\%$ and $42.5\%$ ones, and the one seed that leaves the band is the seed that does not fit the trained depths.

\begin{table}[htbp]
\centering
\small
\caption{Share of predicted ones in the decoded row under $R_{12}$ (five model seeds per family, final test, the checkpoints of Table~\ref{tab:abl-main}). Depth 60 lists the seeds in increasing order, or their range when they agree to within 2 points. The departure is the largest distance from $50\%$ over depths 13--60, as a range over seeds. The last column counts seeds that leave the $25$--$75\%$ band at any of those depths. Two CoTFormer seeds do not fit the trained depths, so its row mixes a failure to fit with failure at depth. All zeros is a fixed point of Rule~30, so a seed that settles there has trivially consistent transitions.}
\label{tab:ones-rate}
\begin{tabular}{@{}lccc@{}}
\toprule
Model & Ones at depth 60 (\%) & Departure from 50\% & Leaving 25--75\% \\
\midrule
LSTM-UT & 49.8--50.0 & 0.3--0.4 & 0/5 \\
Linear memory & 50.0 & 0.3 & 0/5 \\
Decay BUT, \(\lambda=0.9\) & 50.0 & 0.3 & 0/5 \\
Decay BUT, \(\lambda=0.5\) & 50.0 & 0.3 & 0/5 \\
FPRM BUT & 50.0 & 0.3 & 0/5 \\
FPRM, no re-injection & 50.0 & 0.3 & 0/5 \\
No forget gate & 49.8--51.4 & 0.3--13.2 & 0/5 \\
Gated decay BUT, \(\lambda=0.9\) & 16.9, 42.5, 49.3, 50.0, 50.0 & 0.3--33.6 & 1/5 \\
\midrule
Gated BUT & 0.0, 6.7, 48.6, 84.4, 100.0 & 28.0--50.0 & 5/5 \\
BUT + injection & 0.0, 50.1, 58.9, 78.7, 92.1 & 16.0--50.0 & 3/5 \\
BUT & 7.3, 11.6, 45.9, 93.1, 93.4 & 26.7--49.8 & 5/5 \\
CoTFormer & 0.0, 0.0, 73.8, 76.5, 87.2 & 23.8--50.0 & 4/5 \\
\bottomrule
\end{tabular}
\end{table}

\subsection{Decay alone is sufficient: decay BUT}
\label{app:abl-decay}
The comparison above is across architectures. Decay BUT is an intervention on BUT itself: \(\lambda\) is a fixed constant, not
learned, and it is the only difference. Training, block, parameter count and checkpoint selection are identical
(Table~\ref{tab:app-abl-decay}).

\begin{table}[htbp]
\centering
\small
\caption{Decay BUT against BUT under \(R_{12}\). The leading rate is the shift-quotiented leading Lyapunov rate per repeat
(Section~\ref{app:abl-contraction}) over repeats 31--60, mean \(\pm\) sample SD over five seeds.}
\label{tab:app-abl-decay}
\begin{tabular}{@{}lccc@{}}
\toprule
 & BUT (\(\lambda=1\)) & Decay BUT, \(\lambda=0.9\) & Decay BUT, \(\lambda=0.5\) \\
\midrule
Exact rows at depth 60 (\%) & \(0.00\pm0.00\) & \(99.56\pm0.53\) & \(99.98\pm0.04\) \\
Reach per seed & 23, 18, 24, 19, 31 & 60 (all) & 60 (all) \\
Leading rate, repeats 31--60 & \(+0.039\pm0.055\) & \(-0.0975\pm0.0072\) & \(-0.480\pm0.094\) \\
\(\ln\lambda\) & 0 & \(-0.105\) & \(-0.693\) \\
\bottomrule
\end{tabular}
\end{table}

Keeping \(90\%\) of the stream per repeat is enough to turn failure at depths 18--31 into extrapolation to depth 60 in every seed. The
model with \(\lambda=0.9\) does show a small slope by depth 60 (\(99.56\%\) exact rows; seeds 0 and 2 at \(99\%\)).

\paragraph{The decay acts exactly as written.}
A per-token constant offset of \(x\) is invisible to the block, because every LayerNorm removes it, so it is carried only by the
\(\lambda x\) term. Its measured per-repeat rate is 0 for BUT (within \(10^{-6}\)), so the offset is carried forever; \(-0.105361\) for
decay BUT with \(\lambda=0.9\), against \(\ln0.9=-0.105361\), at every repeat, row and seed; and \(-0.693147\) for \(\lambda=0.5\)
(range \(-0.694\) to \(-0.689\) while the offset is still above \(10^{-10}\) of its starting size, after which it reaches float64
precision). No decoded bit ever changes. This is a check that the model and the measurement do what the equation says, not a
scientific result.

\paragraph{Almost all of the contraction at \(\lambda=0.9\) is \(\lambda\) itself.}
The leading Lyapunov rate cannot be below \(\ln\lambda\), since the offset direction alone has rate \(\ln\lambda\). At \(\lambda=0.9\)
the measured rate is \(-0.0975\), so the trained block adds only \(+0.008\) on top of \(\ln\lambda\), less than BUT's block on its own
(\(+0.04\)). All five seeds lie between \(-0.086\) and \(-0.104\). At \(\lambda=0.5\) the block adds more (\(+0.21\)), but the rate
stays strongly negative. Inside the trained depths the rate at \(\lambda=0.9\) is weaker (\(-0.061\) over repeats 1--12) than beyond
them (about \(-0.10\)).

This shows that a fixed decay of the carried state is \emph{sufficient} for depth extrapolation on this task; nothing else in BUT has
to change. It does not show that decay is \emph{necessary}: that rests on the cross-family comparison of
Section~\ref{app:abl-ingredient}, which is a correlation: no undecayed-stream seed reaches depth 60, but two of the five no-forget seeds, which have no decaying state, do. The trained weights differ between BUT
and decay BUT, so the effect of \(\lambda\) includes its effect on what training learns. We have not tested whether some \(\lambda\)
close to 1, for example 0.98 or 0.99, still fails.

\subsection{Contraction}
\label{app:abl-contraction}

\paragraph{Local error growth.}
We report the leading shift-quotiented Lyapunov rate: the per-repeat log growth of the fastest-growing small perturbation, after
removing per-token offsets, which the model cannot see (Section~\ref{app:abl-decay}). Table~\ref{tab:app-abl-lyapunov} gives two windows.

\begin{table}[htbp]
\centering
\small
\caption{Leading shift-quotiented Lyapunov rate per repeat (mean \(\pm\) sample SD over five seeds) inside the trained depths
(repeats 1--12) and far beyond them (repeats 31--60). Exact rows at depth 60 are on the 64 diagnostic rows, not the 4,096-row test
split.}
\label{tab:app-abl-lyapunov}
\begin{tabular}{@{}lrrr@{}}
\toprule
Model & Repeats 1--12 & Repeats 31--60 & Exact rows at 60 (\%) \\
\midrule
LSTM-UT & \(-0.123\pm0.046\) & \(-0.039\pm0.032\) & 99.4 \\
Linear read & \(-0.112\pm0.039\) & \(-0.046\pm0.019\) & 99.4 \\
Linear memory & \(-0.143\pm0.085\) & \(-0.082\pm0.048\) & 100 \\
Decay BUT, \(\lambda=0.9\) & \(-0.061\pm0.028\) & \(-0.098\pm0.007\) & 100 \\
Decay BUT, \(\lambda=0.5\) & \(-0.410\pm0.106\) & \(-0.480\pm0.094\) & 100 \\
FPRM BUT & \(-0.933\pm0.154\) & \(-0.999\pm0.134\) & 100 \\
FPRM, no re-injection & \(-0.798\pm0.188\) & \(-0.899\pm0.140\) & 100 \\
No forget gate & \(+0.013\pm0.008\) & \(+0.045\pm0.046\) & 43.1 \\
Gated decay BUT, \(\lambda=0.9\) & \(-0.021\pm0.080\) & \(+0.019\pm0.113\) & 41.3 \\
Gated BUT & \(+0.008\pm0.013\) & \(+0.013\pm0.012\) & 0 \\
BUT + injection & \(+0.033\pm0.025\) & \(+0.028\pm0.030\) & 0 \\
BUT & \(+0.021\pm0.026\) & \(+0.039\pm0.055\) & 0 \\
\bottomrule
\end{tabular}
\end{table}

Every family that extrapolates in every seed has a negative mean rate in every window. Every undecayed-stream family and the no-forget model has a positive one in windows 1--12,
21--30 and 31--60; over repeats 13--20, BUT and BUT with injection are slightly negative (\(-0.004\) and \(-0.007\)). Per seed the sign
does not separate perfectly: over repeats 31--60 one LSTM-UT seed has \(+0.003\) and is exact at depth 60, and two BUT seeds have
\(-0.003\) and \(-0.0002\) and fail. A rate this close to zero says little either way. FPRM does not need re-injection to contract
(\(-0.90\) against \(-1.00\)).

Gated decay BUT fits neither pattern (Table~\ref{tab:app-abl-gateddecay}). Over repeats 13--20 its four fitted seeds have rates of \(-0.082\) to
\(-0.101\), close to \(\ln0.9=-0.105\), and halve a small rotation about as fast as decay BUT. Seeds 1 and 4 stay negative and reach depth 60.
Seeds 2 and 3 stop at depth 24 and have late rates of \(+0.101\) and \(+0.173\). In both, the per-repeat growth averaged over the 64 rows weakens from
depth 21, is still negative at depth 24 (\(-0.073\) and \(-0.033\)), where \(94\%\) of those rows are exact, and stays positive from depths 32 and 29 on, where \(75\%\) and \(88\%\) are.
The average mixes rows that are still exact with rows that have gone wrong, so it does not show which comes first. What it does show is that a rate near
\(\ln\lambda\) while the rows are exact does not guarantee that they stay exact, and that the late rate of Table~\ref{tab:abl-main} records
these failures and does not predict them. The stream norm changes by under \(10\%\) from depth 12 to depth 42, and the rows do not saturate (Table~\ref{tab:ones-rate}).
Seed 4 has a positive rate within the trained depths (\(+0.032\)), a negative one beyond them, and reaches depth 60. We do not know why seeds 2 and 3 fail.

\begin{table}[htbp]
\centering
\small
\caption{Gated decay BUT per seed (\(R_{12}\)). \emph{Depth 60} and \emph{Reach}: as in Table~\ref{tab:app-abl-extrapolation}. The four rate columns are the leading
shift-quotiented Lyapunov rate per repeat in each window of repeats. \emph{Rotation}: per-repeat log rate of a \(0.01\)\,rad rotation over repeats 13--20, as in
Table~\ref{tab:app-abl-halving}, on 64 exact rows (26 for seed 0). \emph{Recovery}: as in Table~\ref{tab:abl-main}.}
\label{tab:app-abl-gateddecay}
\begin{tabular}{@{}crcrrrrrr@{}}
\toprule
Seed & Depth 60 & Reach & 1--12 & 13--20 & 21--30 & 31--60 & Rotation & Recovery \\
\midrule
0 & \(0.00\) & -- & \(+0.085\) & \(-0.020\) & \(+0.028\) & \(-0.020\) & \(-0.062\) & 0.0 \\
1 & \(100.00\) & 60 & \(-0.098\) & \(-0.101\) & \(-0.098\) & \(-0.089\) & \(-0.109\) & 77.3 \\
2 & \(8.18\) & 24 & \(-0.096\) & \(-0.095\) & \(-0.048\) & \(+0.101\) & \(-0.102\) & 24.9 \\
3 & \(2.34\) & 24 & \(-0.027\) & \(-0.082\) & \(-0.014\) & \(+0.173\) & \(-0.092\) & 12.5 \\
4 & \(95.73\) & 60 & \(+0.032\) & \(-0.091\) & \(-0.084\) & \(-0.069\) & \(-0.101\) & 56.3 \\
\bottomrule
\end{tabular}
\end{table}

The shift quotient matters for the cell models. In LSTM-UT, the linear read and linear memory, the raw leading direction is almost
entirely a per-token offset (shift energy \(88.6\%\), \(98.8\%\) and \(99.8\%\) at repeat 60), and the raw rate is near zero
(\(-0.002\), \(-0.0003\), \(-0.0006\)), so only the shift-quotiented rate is meaningful there. In decay BUT and FPRM the offset itself
decays, so raw and quotiented rates coincide (shift energy \(0.3\)--\(3.4\%\)). Two caveats apply. Without the forget gate the cell is
an integrator, so the cell-to-cell block of the Jacobian is the identity and the leading rate sits near zero by construction; it is not
comparable with the others, although its late window still separates seeds (\(+0.001\) and \(+0.003\) for the two seeds that reach 60,
\(+0.038\) to \(+0.105\) for the three that fail). FPRM's learned scales differ per channel, so a per-token offset does not stay an
offset: after one repeat about \(60\%\) of it is visible to LayerNorm, and at 1.0\,rad it flips up to \(31\%\) of bits in one seed. For
FPRM the shift quotient is therefore not the removal of a symmetry, but because its raw and quotiented rates coincide, no number
changes.

\paragraph{Halving time of a small perturbation.}
We rotate \(x\) by 0.01\,rad at repeat 13 and measure the per-repeat rate over repeats 13--20, on rows the model still gets exactly
right through repeat 20. The rates are the same at 0.001 and 0.1\,rad, so this is the small-perturbation regime
(Table~\ref{tab:app-abl-halving}).

\begin{table}[htbp]
\centering
\small
\caption{Contraction of a 0.01\,rad rotation of \(x\) at repeat 13, over repeats 13--20: per-repeat log rate (mean \(\pm\) sample SD
over five seeds) and the number of repeats to halve the perturbation.}
\label{tab:app-abl-halving}
\begin{tabular}{@{}lrr@{}}
\toprule
Model & Rate & Halving (repeats) \\
\midrule
FPRM BUT & \(-1.168\pm0.123\) & 1 \\
FPRM, no re-injection & \(-1.022\pm0.162\) & 1 \\
Decay BUT, \(\lambda=0.5\) & \(-0.604\pm0.054\) & 1 \\
Linear memory & \(-0.209\pm0.063\) & 3 \\
Linear read & \(-0.167\pm0.022\) & 4 \\
LSTM-UT & \(-0.163\pm0.039\) & 4 \\
No forget gate & \(-0.140\pm0.050\) & 5 \\
Decay BUT, \(\lambda=0.9\) & \(-0.118\pm0.003\) & 6 \\
Gated decay BUT, \(\lambda=0.9\) & \(-0.093\pm0.018\) & 7 \\
BUT + injection & \(-0.022\pm0.005\) & 32 \\
BUT & \(-0.018\pm0.005\) & 39 \\
Gated BUT & \(-0.007\pm0.006\) & 95 \\
\bottomrule
\end{tabular}
\end{table}

Every family that extrapolates in every seed halves an error in at most 6 repeats, and every undecayed-stream family
takes at least 32, a sharper separation than the Lyapunov sign. A short halving time is not sufficient, however: the no-forget model (5 repeats) and
gated decay BUT (7) each reach depth 60 in two of five seeds. Gated decay BUT has no cell to hold an error, and we have no explanation for it.

In the no-forget model \(x\) contracts as fast as LSTM-UT's (\(-0.140\)) although the model is unreliable: its cell keeps
the error. The rotation moves only \(x\), so it reaches the cell one repeat later. From there to repeat 60 the cell gap between the
perturbed and clean branches (Table~\ref{tab:app-abl-cellgap}) shrinks in every seed with a forget gate (LSTM-UT by 3--490 times, linear read by 31--113 times) and grows
in every no-forget seed (3--4 times in the seeds with flat or slowly growing error, 13 and 190 times in the other two). Relative to the
clean state at repeat 60, the two no-forget seeds that reach depth 60 have \(x\) error \(2\)--\(9\times10^{-4}\), as in LSTM-UT
(\(0.3\)--\(11\times10^{-4}\)), and cell error \(1.9\times10^{-4}\); two failing seeds have \(x\) error \(1.6\times10^{-2}\) and
\(5.0\times10^{-2}\) with cell error \(0.9\) and \(14\times10^{-3}\); the third failing seed has a flat \(x\) error of
\(2.4\times10^{-3}\). Fast contraction of \(x\) alone is therefore not enough: every state that is carried forward has to forget. In
linear memory the cell gap at repeat 60 is \(0.6\times10^{-6}\) to \(1.1\times10^{-4}\) of the clean cell.

\begin{table}[htbp]
\centering
\footnotesize
\caption{Per-seed evidence that the cell keeps the error without a forget gate (\(R_{12}\)). \emph{Reach}: as in
Table~\ref{tab:abl-main}. \emph{Late rate}: leading Lyapunov rate per repeat over repeats 31--60, whose mean over seeds is the
\emph{Rate} of Table~\ref{tab:abl-main}. \emph{Depth 1,000}: exact rows (\%) as in Table~\ref{tab:depth1000}.
\emph{\(|c|\), 60}: mean absolute cell value at repeat 60. The remaining columns follow a \(0.01\)\,rad rotation of \(x\) at
repeat 13 (medians over branches). \emph{Rate of \(x\)}: per-repeat log rate of the centred \(x\) distance over repeats 13--20.
\emph{Cell gap}: distance between the perturbed and clean cells at repeats 14 and 60, and its change. \emph{\(x\) error}:
\(x\) gap at repeat 60 relative to the clean \(x\) norm.}
\label{tab:app-abl-cellgap}
\setlength{\tabcolsep}{3.5pt}
\resizebox{\textwidth}{!}{\begin{tabular}{@{}lcrrrrrrrlr@{}}
\toprule
Model & Seed & Reach & Late rate & Depth 1,000 & \(|c|\), 60 & Rate of \(x\) & Cell gap, 14 & Cell gap, 60 & Change & \(x\) error, 60 \\
\midrule
LSTM-UT & 0 & 60 & \(-0.024\) & \(94.59\) & \(0.9\) & \(-0.128\) & \(0.238\) & \(0.083\) & shrinks \(2.9\times\) & \(8.0\times10^{-4}\) \\
 & 1 & 60 & \(-0.044\) & \(93.72\) & \(0.6\) & \(-0.153\) & \(0.194\) & \(0.004\) & shrinks \(43\times\) & \(3.3\times10^{-4}\) \\
 & 2 & 60 & \(-0.048\) & \(68.20\) & \(0.8\) & \(-0.144\) & \(0.264\) & \(0.007\) & shrinks \(39\times\) & \(2.6\times10^{-4}\) \\
 & 3 & 60 & \(+0.003\) & \(4.74\) & \(2.4\) & \(-0.161\) & \(0.287\) & \(0.077\) & shrinks \(3.7\times\) & \(1.1\times10^{-3}\) \\
 & 4 & 60 & \(-0.083\) & \(99.98\) & \(0.6\) & \(-0.228\) & \(0.225\) & \(4.6\times10^{-4}\) & shrinks \(485\times\) & \(3.4\times10^{-5}\) \\
\midrule
No forget gate & 0 & 35 & \(+0.038\) & \(0.00\) & \(34.5\) & \(-0.188\) & \(0.170\) & \(0.672\) & grows \(4.0\times\) & \(2.4\times10^{-3}\) \\
 & 1 & 60 & \(+0.001\) & \(99.81\) & \(40.0\) & \(-0.156\) & \(0.148\) & \(0.525\) & grows \(3.5\times\) & \(2.3\times10^{-4}\) \\
 & 2 & 27 & \(+0.078\) & \(0.00\) & \(37.3\) & \(-0.160\) & \(0.176\) & \(2.35\) & grows \(13\times\) & \(1.6\times10^{-2}\) \\
 & 3 & 60 & \(+0.003\) & \(0.00\) & \(37.4\) & \(-0.139\) & \(0.147\) & \(0.490\) & grows \(3.3\times\) & \(9.2\times10^{-4}\) \\
 & 4 & 19 & \(+0.105\) & \(0.00\) & \(31.1\) & \(-0.057\) & \(0.171\) & \(32.50\) & grows \(190\times\) & \(5.0\times10^{-2}\) \\
\bottomrule
\end{tabular}}
\end{table}

\paragraph{Neutral seeds.}
The two no-forget seeds that reach depth 60 are not contracting: their late leading rates are \(+0.001\) and \(+0.003\), against
\(+0.038\) to \(+0.105\) in the three that fail (Table~\ref{tab:app-abl-cellgap}). A neutral rollout is not a stable one. One of
the two falls below \(50\%\) of rows by depth 134, and the one LSTM-UT seed whose late rate is also about zero (\(+0.003\), with a
cell-to-cell rate of \(-0.004\) against \(-0.05\) to \(-0.46\) in the others) is the seed that falls below \(50\%\) by depth 491.
The four LSTM-UT seeds with negative late rates all keep at least \(68\%\) of rows at depth 1,000. Within that group the rate does
not rank the seeds, and with three neutral seeds we cannot say how often a neutral rollout lasts. Without a forget gate the cell
also saturates: its mean absolute value is \(7\)--\(8\) at depth 12 and \(31\)--\(40\) at depth 60 in every seed, against
\(0.6\)--\(2.4\) in LSTM-UT, so the \(\tanh\) read has become a fixed sign pattern passed through the output gate.

\paragraph{Interpretation.}
\citet{lu2018attractor} give the right frame. A recurrent model trained on short horizons and then iterated reproduces a system's
long-run behaviour only if the set of states that encode valid trajectories is stable when the loop is closed: perturbations
\emph{transverse} to it must shrink. Fitting the trained depths does not ask for this beyond depth 12, which is why every model fits
and only some extrapolate. Rule~30 rows are discrete, so an infinitesimal perturbation can never move the state to another valid row:
every small perturbation is transverse, and a negative leading rate is exactly this stability.

A weight-1 stream with nothing else to contract it is the obstacle. For BUT the linearisation is \(I+J_\Delta\), so every direction the block cannot see keeps
multiplier exactly 1; for a leaky integrator \citep{yildiz2012echo} it is \(\lambda I+J_\Delta\). Decay BUT is that leaky integrator.
The model does not forget its input in the echo-state sense (the input row is the initial state, and forgetting it would be failure);
it contracts transversely to the set of valid row codes while the block keeps rewriting the row. In FPRM without re-injection the row
survives 60 repeats of \(\times0.14\) decay only because the block rewrites it every repeat.

\subsection{Faithful computation is the best a rollout can do after an error}
\label{app:abl-faithful}
Rule~30 does not forgive a wrong bit. It is left-permutive: the new cell \(i+1\) is \(s_i\oplus(s_{i+1}\lor s_{i+2})\), so a difference
in the rightmost wrong cell always changes its right neighbour at the next step. On an infinite line two rows that differ can therefore
never become equal again. On our 64-cell ring, none of 4,096 single-bit flips merged back within 60 steps, and the number of differing
bits grew to 32.0 of 64, which is chance.

So once a row is wrong, no model that applies Rule~30 to its own output can return to the true row. Without an external anchor, the
best possible behaviour after an error is to keep evolving the wrong row faithfully. We call a repeat \emph{faithful} when every bit
equals Rule~30 applied to the model's own previous row, and the row is neither constant nor frozen: the share of ones and the share of
changed bits both lie in \([0.25,0.75]\). A random row passes by chance with probability about \(2^{-64}\). In the terms of
\citet{lu2018attractor}, faithful computation on a wrong row is a return to the invariant set at a different point of it: the model
loses the forecast and keeps the climate.

\subsection{The read is needed at every step: closing the output gate}
\label{app:abl-gate}
This experiment tests causally what the cell's read does during the computation. It applies to the four models with a read:
LSTM-UT, linear read, no forget gate and linear memory (20 seeds). Models without a cell have no read to skip.

\paragraph{Intervention.}
One repeat of these models is, in order: (1) \(x\leftarrow x+\text{attention}+\text{MLP}\); (2) the cell is updated from that \(x\);
(3) the read adds the cell to \(x\) (\(o\odot\tanh(c)\), \(o\odot c\), or \(c\) for linear memory). Closing the gate skips step 3 for
one or more repeats while steps 1 and 2 still run. It is a switch in the model's own forward pass, never set in training, not a change
to the weights. We close it at repeat 6, 13, 20 or 30 for 1--15 repeats (or never reopen it), then run the unmodified model to repeat
60. In \emph{update} mode the cell keeps being written during the closure; in \emph{frozen} mode it is held at its value at the start.
For example, for a closure at repeat 13 lasting one repeat, repeats 1--13 are the unmodified model, repeat 14 runs attention, MLP and
the cell update but skips the read, and repeats 15--60 are the unmodified model again. Every repeat is decoded.
\paragraph{One skipped read breaks the computation, in every seed.}
After a single skipped read, no row is ever exactly right again: exact-row accuracy at repeat 60 is \(0\%\) in all 20 seeds, for every
start depth, cell mode and closure length, and no row is exact at any repeat from two after reopening on. This is not because the
skipped repeat itself is wrong, and not because the cell is lost. The skipped repeat still decodes almost correctly: its cell accuracy
is \(93\)--\(100\%\) in LSTM-UT, \(76\)--\(99\%\) in the linear read, \(88\)--\(100\%\) without the forget gate, and \(100\%\) in every linear-memory seed
(\(100\%\) exact rows in all five). Its margin collapses, though: in LSTM-UT and the linear read the minimum logit margin falls from
9.0--20.4 to between \(-7.2\) and \(+5.9\). The next, unmodified repeat is wrong: one repeat after reopening, cell accuracy is
\(56\)--\(75\%\) in LSTM-UT, \(53\)--\(87\%\) in the linear read, \(58\)--\(72\%\) without the forget gate and \(72\)--\(87\%\) in linear memory. The cell is
intact: in update mode the cell after the closed repeat equals the clean cell exactly, because it is written before the read from an
input that is still clean (measured distance 0.00 in LSTM-UT). What breaks the computation is the missing term in \(x\), not missing
memory.

The read is a large part of each repeat's update: skipping it moves \(x\) by 0.57--0.73 of its centred norm in LSTM-UT (an angle of
0.61--0.79\,rad), and by 0.44--0.50 (0.42--0.51\,rad) in linear memory. The read is therefore used at every repeat, and the state that
decodes correctly on the closed repeat cannot drive the next one. This is the same readable-but-not-reusable pattern as in
Section~\ref{app:abl-recovery}, produced here by removing one specific term.

\paragraph{Reopening the gate restores \emph{faithful} computation.}
By Section~\ref{app:abl-faithful} the original row can never come back. The question is whether the model returns to applying Rule~30
to its own output once the read is restored. After a one-repeat closure at repeat 13 in update mode, the mean self-transition accuracy
rises again over the next repeats, on rows with a normal share of ones (\(37\)--\(53\%\) six repeats after reopening in 19 of 20 seeds; one
LSTM-UT seed has \(12\%\)). It reaches at least \(95\%\) within 12 open repeats in 13 of 20 seeds: two LSTM-UT seeds, three linear-read seeds,
four no-forget seeds and four linear-memory seeds. In five of these seeds it is exactly \(100\%\) within 12 open repeats, and in two more by
repeat 60.

\begin{table}[htbp]
\centering
\small
\caption{Share of branches (\%) ending as a faithful evolution of a wrong row (repeats 51--60) after a one-repeat gate closure, update
mode, pooled over closures at repeats 6, 13, 20 and 30. Frozen mode differs by at most 6 points, except one linear-memory seed (33
against 20).}
\label{tab:app-abl-gate}
\begin{tabular}{@{}lrrrrrc@{}}
\toprule
Model & Seed 0 & Seed 1 & Seed 2 & Seed 3 & Seed 4 & Seeds \(\geq50\%\) \\
\midrule
LSTM-UT & 2 & 100 & 0 & 0 & 0 & 1/5 \\
Linear read & 99 & 1 & 2 & 58 & 0 & 2/5 \\
No forget gate & 5 & 97 & 20 & 97 & 0 & 2/5 \\
Linear memory & 0 & 100 & 33 & 95 & 88 & 3/5 \\
\bottomrule
\end{tabular}
\end{table}

Whether the restored computation lasts to repeat 60 is seed dependent (Table~\ref{tab:app-abl-gate}). The seeds that do not stay faithful
end frozen or as all-zero rows (an all-zero row is a fixed point of Rule~30 and is excluded by the definition): for example, LSTM-UT
seed 2 ends frozen in \(100\%\) of branches and seed 3 all-zero in \(59\%\). In the seeds that recover, the restored computation is
robust to longer closures. LSTM-UT seed 1 with the cell held fixed stays faithful in \(62\)--\(92\%\) of branches for closures of 5--15
repeats, and two linear-memory seeds stay faithful in \(39\)--\(50\%\) of branches even after 15 skipped reads. With the gate never
reopened, no branch of any seed is faithful at the end.

\paragraph{Closing the gate is not the same as a random rotation of the same size.}
A one-repeat closure moves \(x\) by about the angle of a 0.6\,rad rotation. In LSTM-UT and the linear read, a random rotation of that
size leaves many more branches faithful at the end than the closure does; the no-forget model is the exception
(Table~\ref{tab:app-abl-rotation-closure}). Closing the gate removes a specific, learned term, and in LSTM-UT and the linear read that term
carries much of the decision margin. In the no-forget model the margin barely changes (6.1--15.5 clean, 6.4--15.9 closed), and there
the closure is no more damaging than a rotation.

\begin{table}[htbp]
\centering
\small
\caption{Share of all branches (\%) ending as a faithful evolution of a wrong row after a 0.6\,rad random rotation of \(x\) at repeat
13 (Rot.) or a one-repeat gate closure in update mode, pooled over closures at repeats 6, 13, 20 and 30 (Close).}
\label{tab:app-abl-rotation-closure}
\begin{tabular}{@{}lrrrrrr@{}}
\toprule
 & \multicolumn{2}{c}{LSTM-UT} & \multicolumn{2}{c}{Linear read} & \multicolumn{2}{c}{No forget gate} \\
\cmidrule(lr){2-3}\cmidrule(lr){4-5}\cmidrule(l){6-7}
Seed & Rot. & Close & Rot. & Close & Rot. & Close \\
\midrule
0 & 39 & 2 & 94 & 99 & 6 & 5 \\
1 & 99 & 100 & 25 & 1 & 68 & 97 \\
2 & 22 & 0 & 43 & 2 & 20 & 20 \\
3 & 29 & 0 & 80 & 58 & 48 & 97 \\
4 & 24 & 0 & 8 & 0 & 0 & 0 \\
\bottomrule
\end{tabular}
\end{table}

\paragraph{What this does and does not show.}
The read is causally necessary at every repeat in all 20 seeds of all four models with a read: one skipped read loses the row
permanently, although the skipped repeat itself decodes almost correctly and the cell is intact. Restoring the read restores valid
Rule~30 computation on the now wrong row within a few repeats in most seeds, and permanently in 8 of 20. The computation is carried by
the loop through the read, not stored in the cell as a recoverable copy of the row. The experiment does not show that the cell is a
store from which the row can be restored: the original row never returns, and by Section~\ref{app:abl-faithful} it cannot. Nor does it
show that lasting recovery is a property of the architecture, since it occurs in 1 of 5 LSTM-UT seeds. It concerns the extrapolation
task only, not delayed recall.

\subsection{Recovery after damage}
\label{app:abl-recovery}
We rotate every token's \(x\) once, by 0.6 or 1.0\,rad in a random direction, at repeat 6 or 13, and let the model run freely to repeat
60. There are 512 branches per seed (4 directions \(\times\) 2 signs \(\times\) 64 rows).

\begin{table}[htbp]
\centering
\small
\caption{Recovery to faithful computation after a rotation of \(x\): percentage of broken branches that return to at least five
consecutive faithful repeats within 15 repeats, pooled over five seeds. Only rows on which the unperturbed model stays faithful
throughout that window count, so a failure to recover cannot be the model's own collapse. The number of broken branches is in
brackets where they are few.}
\label{tab:app-abl-recovery}
\begin{tabular}{@{}lrrrr@{}}
\toprule
 & \multicolumn{2}{c}{0.6\,rad} & \multicolumn{2}{c}{1.0\,rad} \\
\cmidrule(lr){2-3}\cmidrule(l){4-5}
Model & at 6 & at 13 & at 6 & at 13 \\
\midrule
FPRM BUT & 94.0 & 93.5 & 57.3 & 58.2 \\
FPRM, no re-injection & 88.5 & 89.5 & 54.7 & 56.7 \\
Decay BUT, \(\lambda=0.5\) & 81.8 & 80.4 & 32.6 & 34.1 \\
No forget gate & 80.4 & 93.6 & 0.2 & 0.2 \\
Linear read & 74.6 & 68.7 & 4.5 & 5.0 \\
LSTM-UT & 63.1 & 56.0 & 14.5 & 13.7 \\
Linear memory & 55.2 & 50.2 & 7.5 & 7.5 \\
Decay BUT, \(\lambda=0.9\) & 52.2 & 41.7 & 0.8 & 0.2 \\
Gated decay BUT, \(\lambda=0.9\) & 38.8 & 22.6 (1,886) & 0.0 & 0.0 (1,888) \\
BUT + injection & 10.0 & 2.1 (952) & 0 & 0 (952) \\
BUT & 7.7 & 1.1 (1,056) & 0 & 0 (1,056) \\
Gated BUT & 0.3 & 0.0 (288) & 0 & 0 (288) \\
\bottomrule
\end{tabular}
\end{table}

Per seed at 0.6\,rad and repeat 6, recovery is \(38\)--\(97\%\) for LSTM-UT, \(48\)--\(90\%\) for the linear read, \(12\)--\(98\%\)
for linear memory, \(13\)--\(88\%\) for decay BUT with \(\lambda=0.9\), \(42\)--\(98\%\) with \(\lambda=0.5\), \(0\)--\(77\%\) for gated decay BUT, \(90\)--\(97\%\) for
FPRM and \(67\)--\(99\%\) for FPRM without re-injection. At 1.0\,rad the cell models recover almost only in one seed each (one seed each
for LSTM-UT and the linear read, two for linear memory). Each of the seven decaying families that reach depth 60 recovers more often than every undecayed-stream
family at 0.6\,rad (\(42\)--\(94\%\) against \(0\)--\(10\%\); Table~\ref{tab:app-abl-recovery}); gated decay BUT is between, at \(23\)--\(39\%\). Among the stream-decay models, faster
contraction goes with recovery from larger damage: at 1.0\,rad FPRM (\(55\)--\(58\%\)) exceeds decay BUT with \(\lambda=0.5\)
(\(33\)--\(34\%\)), which exceeds \(\lambda=0.9\) (\(0.2\)--\(0.8\%\)). Among the cell models it does not: linear memory contracts
fastest (halving time 3) and recovers least. A negative rate describes a neighbourhood of the valid codes, and a 1.0\,rad rotation
leaves it; local contraction need not predict global return \citep{yildiz2012echo}.

\paragraph{Still faithful at the end.}
We also count the branches that have gone wrong and are still faithful on a wrong row at repeat 60, meaning a faithful streak of at
least 10 repeats ending there, after a 0.6\,rad rotation at repeat 6. Over seeds this share is \(30\)--\(99\%\) for LSTM-UT,
\(26\)--\(92\%\) for the linear read, \(8\)--\(100\%\) for linear memory, \(10\)--\(98\%\) and \(26\)--\(98\%\) for decay BUT with
\(\lambda=0.9\) and \(0.5\), \(22\)--\(93\%\) for FPRM, \(40\)--\(94\%\) for FPRM without re-injection, \(0\)--\(97\%\) for gated decay BUT, and \(0\)--\(55\%\) without the
forget gate. It is \(0\%\) in every undecayed-stream seed, and no such branch is faithful for more than 19 consecutive repeats anywhere
(BUT 15, gated BUT 7, BUT with injection 19). These models fail without any perturbation well before repeat 60 (reach 18--31), so
their \(0\%\) here reflects their own collapse at depth as well as the rotation; the matched-window recovery of
Table~\ref{tab:app-abl-recovery}, which counts only rows the unperturbed model keeps faithful, is the fair comparison. Recovery is not reliability. The no-forget model recovers often, but its failing seeds do
not stay faithful: in one of them \(92\%\) of branches reach a faithful streak of at least 10 repeats, yet only \(5\%\) are still
faithful at repeat 60. In the two seeds that reach depth 60, \(55\%\) and \(46\%\) are.

\paragraph{A row that decodes correctly is not necessarily reusable.}
Right after a 0.6\,rad rotation at repeat 6, the perturbed state still decodes to the exact correct row in at least \(99.6\%\) of
branches in every model but gated decay BUT (\(96.9\%\); \(100\%\) in its four fitted seeds). Yet, on rows the unperturbed model keeps exact for 15 more repeats, a later repeat is wrong in a large share
of branches (Table~\ref{tab:app-abl-readable}). The decoded row is readable, but the state that produced it does not drive the next steps
correctly. \citet{zhou2026retain} draw the same distinction between readable answers and reusable states in looped Transformers,
using interventions that freeze a position's state.

\begin{table}[htbp]
\centering
\small
\caption{Branches (\%) in which a later repeat is wrong after a 0.6\,rad rotation of \(x\) that still decodes to the exact correct
row, on rows the unperturbed model keeps exact for 15 more repeats, pooled over seeds.}
\label{tab:app-abl-readable}
\begin{tabular}{@{}lrr@{}}
\toprule
Model & Rotation at 6 & Rotation at 13 \\
\midrule
FPRM BUT & 56.4 & 57.9 \\
No forget gate & 66.6 & 69.7 \\
FPRM, no re-injection & 69.4 & 68.4 \\
Decay BUT, \(\lambda=0.5\) & 72.8 & 73.2 \\
Linear read & 77.9 & 82.4 \\
LSTM-UT & 80.4 & 84.3 \\
Linear memory & 93.4 & 95.3 \\
Decay BUT, \(\lambda=0.9\) & 93.9 & 98.7 \\
Gated decay BUT, \(\lambda=0.9\) & 99.5 & 99.9 \\
BUT + injection & 91.8 & 100 \\
BUT & 98.8 & 100 \\
Gated BUT & 100 & 100 \\
\bottomrule
\end{tabular}
\end{table}

\subsection{Summary}
\label{app:abl-summary}
A decaying carried state separates extrapolation from failure in every family but one: all 35 seeds of seven families with one reach depth 60, and
all 25 seeds of the five undecayed-stream families stop at depths 14--31. The exception is gated decay BUT, which contracts while its rows are exact and
reaches depth 60 in two of five seeds. Gates, read saturation, input re-injection and extra
parameters are each neither needed nor sufficient. Decay is sufficient by intervention: multiplying BUT's stream by a fixed 0.9 or 0.5
per repeat, with nothing else changed, turns reach 18--31 into reach 60 in every seed, and at \(\lambda=0.9\) nearly all of the model's
contraction is \(\lambda\) itself. The models that extrapolate contract: every one halves a small error in at most 6 repeats, and every
undecayed-stream model takes at least 32. Every state carried forward has to forget; without the forget gate, \(x\) contracts but the
cell keeps the error. The read is used at every step: one skipped read loses the row in all 20 seeds that have a read, while the
skipped repeat itself still decodes and the cell is intact, and restoring the read restores valid computation on the wrong row within a
few repeats in most seeds and permanently in 8 of 20. After damage, the decaying models return to faithful computation and the
undecayed-stream models essentially never do; by Rule~30's left-permutivity, faithful computation on a wrong row is the best any
rollout can do. Decay BUT, the rotations and the gate closures are interventions; that decay is necessary rests on the
cross-family comparison, with five seeds per family on one task at width 64.

In both cell models with a forget gate, the mean forget gate at depth 60 is 0.18--0.31, and fewer than \(6\%\) of gate entries exceed
0.95 in any seed; linear memory's learned decay is 0.28--0.35 on average, with no channel above 0.95. In these models, which are not trained
to recall, the cell is a fast-decaying side channel used at every step, not a long-term store; in the recall models it holds some earlier
rows (Section~\ref{sec:recall-lookup}).

\section{Language modelling details}
\label{app:lm}
\subsection{Setup}
All models are GPT-2-style causal language models \citep{radford2019language} trained on OpenWebText2 \citep{gao2020pile}, following the
CoTFormer protocol \citep{mohtashami2025cotformer}. Unlike the cellular-automaton models, each has a 2-layer prelude, a 9-layer core
looped \(R\in\{5,8,10\}\) times, and a 1-layer coda (effective depth 48, 75, and 93), with width 384, six heads, rotary positions, and
sequence length 256. Training uses 40,000 AdamW steps at batch size 256 (about 2.6B tokens), learning rate \(10^{-3}\) with 20\% warmup
and cosine decay, and one model seed per configuration. In LSTM-UT the cell persists through the prelude, the loop, and the coda: each
prelude and coda block, and each complete application of the looped core, is followed by one gated update whose previous input is the
block's or core's input and whose proposed input is its output. ``LN-'' variants apply a layer normalisation after every repeat, as in
LN-CoTFormer \citep{mohtashami2025cotformer}. For a parameter-matched comparison at \(R=5\), BUT and CoTFormer were also trained at width
408 (44.51M parameters, against 45.28M for LSTM-UT). We report validation perplexity at step 40,000 and the median time per training step
after step 1,000, comparing times only between runs at the same width and repeat count on the same GPU type.

\subsection{Results}
\begin{figure}[htbp]
\centering
\includegraphics[width=\textwidth]{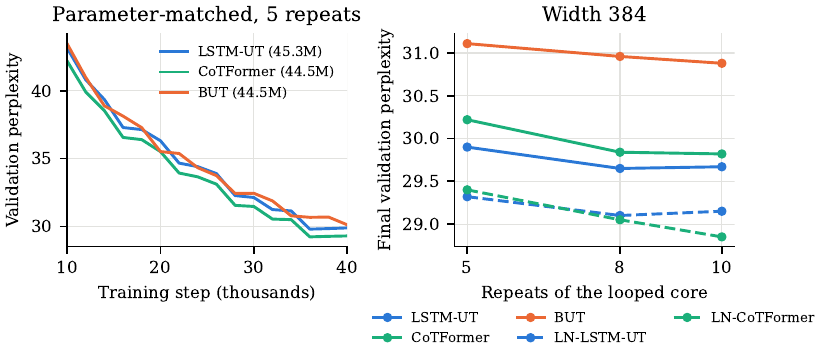}
\caption{OpenWebText2 language modelling. Left: validation perplexity during training for the parameter-matched models with five repeats
(width-408 BUT and CoTFormer, width-384 LSTM-UT). Right: validation perplexity at step 40,000 against the number of repeats at width
384; dashed lines add a layer normalisation after every repeat. One model seed per configuration.}
\label{fig:lm-results}
\end{figure}

Figure~\ref{fig:lm-results} and Table~\ref{tab:lm-results} summarise the runs. At width 384, LSTM-UT reached lower perplexity than BUT
at every repeat count, by 1.2 to 1.3 points, and slightly lower perplexity than CoTFormer (29.90, 29.65, and 29.67 against 30.22, 29.84,
and 29.82). At matched parameter count, however, CoTFormer was better (29.31 against 29.90, with BUT at 30.13), and with a layer
normalisation after every repeat the two were close at \(R=5\) (29.32 for LN-LSTM-UT, 29.40 for LN-CoTFormer) while LN-CoTFormer was
lower at \(R=8\) and \(R=10\).

LSTM-UT was substantially cheaper to train than CoTFormer. On the same GPUs, CoTFormer took \(22.5\%\), \(60.4\%\), and \(90.7\%\) more
time per step than LSTM-UT at \(R=5\), 8, and 10, and \(23.0\%\), \(60.2\%\), and \(89.4\%\) more for the LN variants; total job
wall-clock shows the same ratios. The overhead grows with depth because each CoTFormer repeat attends over keys and values cached from
every earlier repeat, whereas the LSTM-UT cell has a fixed size. LSTM-UT cost \(10.8\%\) to \(13.6\%\) more per step than BUT. With one
seed per configuration, perplexity differences of a few tenths should be treated as indicative. The evidence is that bounded gated memory
is competitive with an expanding cache for language modelling at a fraction of its training cost, not that it improves perplexity over
CoTFormer.

\begin{table}[htbp]
\centering
\scriptsize
\caption{OpenWebText2 validation perplexity at step 40,000 and median time per training step at width 384, one model seed per
configuration. Times at \(R=5\) were measured on A100 GPUs, except for the LN variants; all other times were measured on H200 GPUs.
Bold marks the lowest perplexity in each column.}
\label{tab:lm-results}
\begin{tabular}{@{}lrrrrrrr@{}}
\toprule
& & \multicolumn{3}{c}{Validation perplexity} & \multicolumn{3}{c}{Time per step (ms)} \\
\cmidrule(lr){3-5}\cmidrule(l){6-8}
Model & Parameters & \(R=5\) & \(R=8\) & \(R=10\) & \(R=5\) & \(R=8\) & \(R=10\) \\
\midrule
BUT & 40.56M & 31.11 & 30.96 & 30.88 & 411 & 350 & 427 \\
CoTFormer & 40.56M & 30.22 & 29.84 & 29.82 & 572 & 627 & 902 \\
LSTM-UT & 45.28M & 29.90 & 29.65 & 29.67 & 467 & 391 & 473 \\
\midrule
LN-CoTFormer & 40.56M & 29.40 & \textbf{29.05} & \textbf{28.85} & 331 & 631 & 907 \\
LN-LSTM-UT & 45.28M & \textbf{29.32} & 29.10 & 29.15 & 269 & 394 & 479 \\
\bottomrule
\end{tabular}
\end{table}

\end{document}